\documentclass[11pt,a4paper]{article}
\usepackage{times,latexsym}

\usepackage[T1]{fontenc}
\usepackage[]{tacl2021v1}
\taclpubformattrue

\makeatletter
\def\showauthors@on{T}
\makeatother

\title{GTPO: Stabilizing Group Relative Policy Optimization via \\ Gradient and Entropy Control}

\author{
  Marco Simoni\textsuperscript{\rm 1,3}\thanks{. Equal contribution.},
  Aleksandar Fontana\textsuperscript{\rm 2,3}\footnotemark[1],
  Giulio Rossolini\textsuperscript{\rm 2},
  Andrea Saracino\textsuperscript{\rm 2}, 
  Paolo Mori\textsuperscript{\rm 3} \\
  \textsuperscript{1} National Doctorate on Artificial Intelligence, Sapienza Università di Roma\\
  \textsuperscript{2} Department of Excellence in Robotics and AI, TeCIP, Scuola Superiore Sant’Anna, Pisa \\
  \textsuperscript{3} Institute of Informatics and Telematics, National Research Council of Italy, Pisa \\
}

\usepackage{xspace,mfirstuc,tabulary}
\usepackage{todonotes}

\iftaclpubformat 

\else

\fi

\usepackage{tcolorbox}
\usepackage{pifont}
\usepackage{listings}

\usepackage{times}  
\usepackage{helvet}  
\usepackage{courier}  
\usepackage{graphicx} 
\usepackage{natbib}  
\usepackage{caption} 
\usepackage{algorithm}
\usepackage{algorithmic}
\usepackage{newfloat}
\usepackage{listings}

\usepackage{soul}

\DeclareCaptionStyle{ruled}{labelfont=normalfont,labelsep=colon,strut=off} 
\floatstyle{ruled}
\newfloat{listing}{tb}{lst}{}
\floatname{listing}{Listing}
\usepackage{xcolor}
\usepackage{amsmath}
\usepackage{amssymb}
\usepackage{mathtools}
\usepackage{url}
\usepackage[utf8]{inputenc}
\usepackage[T1]{fontenc}
\usepackage{booktabs}
\usepackage{multirow}
\usepackage{graphicx}
\usepackage{array}
\usepackage[table]{xcolor}

\usepackage{subcaption}  
\usepackage{lipsum}
\newcommand{\NAME}{\textsc{GTPO}}

\newcommand{\cg}{\cellcolor{black!10}}

\newcolumntype{J}{p{0.3cm}@{}}

\usepackage{listings}
\usepackage{upquote}

\lstdefinelanguage{YAML}{
  keywords={true,false,null,yes,no},
  sensitive=false,
  comment=[l]{\#},
  morestring=[b]",
}

\lstdefinestyle{yaml-box}{
  language=YAML,
  basicstyle=\ttfamily\footnotesize,
  frame=single,        
  framerule=0.4pt,     
  columns=fullflexible,
  showstringspaces=false,
  breaklines=true,
  upquote=true
}
\lstdefinestyle{pythonstyle}{
  language=Python,
  basicstyle=\ttfamily\footnotesize,
  keywordstyle=\color{blue},
  commentstyle=\color{gray},
  stringstyle=\color{teal},
  showstringspaces=false,
  numbers=left,
  numberstyle=\tiny\color{gray},
  stepnumber=1,
  breaklines=true,
  frame=single,
  tabsize=2,
  captionpos=b
}

\date{}
\begin{document}
 
\setcounter{secnumdepth}{2} 

\maketitle

\begin{abstract}

Group Relative Policy Optimization (GRPO) is a promising policy-based approach for Large Language Model alignment, yet its performance is often limited by training instability and suboptimal convergence. 
In this paper, we identify and analyze two main GRPO issues: \textit{(i)} the token-level penalization, where valuable tokens shared across different responses receive contradictory feedback signals, leading to conflicting
 gradient updates that can reduce their likelihood; 
and \textit{(ii)} the policy collapse, where negatively rewarded completions
 may penalize confident responses and shift
 model decisions toward unlikely tokens, destabilizing training process.
To address these issues we introduce \textit{GTPO (Group-relative Trajectory-based Policy Optimization)}, which prevents conflicting gradients on valuable tokens
by skipping negative updates while amplifying positive ones and filters out completions whose entropy exceeds a provable threshold, to prevent policy collapse. Unlike GRPO, GTPO does not rely on KL-divergence regularization, eliminating the need for a reference model during training, while still ensuring greater training stability and improved performance, as validated through multiple experiments on GSM8K, MATH, AIME 2024, AIME 2025 and AMC 2023. The Github code is available here\footnote{\url{https://github.com/winstonsmith1897/GTPO}.}.


\end{abstract}

\section{Introduction}

\normalsize

Recent advances in the training of large language models (LLMs) have led to the use of policy-based optimization methods, where the model is treated as a policy that is updated to better match human preferences. Methods such as DPO~\cite{DPO}, RLHF~\cite{rlhf}, and RFT~\cite{RFT} follow this idea: they collect human or heuristic feedback on model outputs and use it as a reward signal to increase the likelihood of preferred or desired responses.
 Group Relative Policy Optimization (GRPO)~\cite{GRPO} is a more recent RL-based method that builds on PPO~\cite{ppo} but removes the need for a separate critic network, i.e., the value function estimator. GRPO compares multiple completions for the same question, each consisting of natural-language reasoning followed by the final answer, and computes advantages from their relative rewards based on completion correctness and formatting quality. The rewards are computed using RL with Verifiable Rewards (RLVR)~\cite{GRPO}, ensuring that they are deterministic and verifiable.
 
 Despite the improvements introduced by GRPO, there are  two key issues identified and analyzed in this work:  
\textit{(i)} undesired \textit{gradient conflicts} affect potentially correct tokens shared across multiple completions of the same group that receive both positive and negative advantages, and  
\textit{(ii)} a \textit{policy collapse} phenomenon, defined as a degradation in policy performance during training, in which negatively rewarded completions may penalize confident responses and flatten tokens likelihood distribution, shifting model decisions toward unlikely tokens.

To address these issues, we propose \textit{Group-relative Trajectory-based Policy Optimization} (\NAME), which treats the sequence of generated tokens (i.e., the completion) as a trajectory of decisions taken by the LLM policy. 
The core idea behind GTPO is to control gradient conflict and entropy explosion while keeping the average output entropy from becoming too low, so that exploration is not prematurely suppressed. To do so, GTPO applies a \textit{conflict-aware gradient correction} mechanism that mitigates gradient conflicts on shared tokens, particularly in the initial and final parts of completions, thus preserving consistency across trajectories.
Furthermore, we show that while the Kullback–Leibler (KL) divergence of GRPO often reacts too slowly in preventing policy collapse, monitoring the average entropy of the output tokens distribution offers a clearer signal of training instability. Building on this, entropy-based regularization terms are applied in \NAME~to control the exploration of trajectories in the same group, also filtering out completions (regardless of their corresponding advantage) whose entropy exceeds a
provable threshold. In this way, GTPO prevents the penalization of important tokens and mitigates policy collapse, improving both the formatting and accuracy of generated completions. Moreover, GTPO removes the KL divergence term used in vanilla GRPO~\cite{GRPO}, avoiding the reference model and making training faster. 
In summary, the contributions of the work are:
\begin{itemize}
    \item We identify two critical issues in GRPO: (i) gradient conflicts on tokens shared across completions within the same group, and (ii) a policy collapse phenomenon that exposes the limitations of the KL term.

    \item We introduce GTPO, a novel policy-based optimization that addresses the previous issues via \textit{conflict-aware gradient corrections} and \textit{entropy control}.


    \item We conduct extensive experiments and ablations studies on mathematical reasoning task to validate the effectiveness of GTPO against GRPO and Supervised fine-tuning (SFT), using  LLaMA 8B~\cite{llama} and Qwen 2.5 (3B)~\cite{qwen}. We demonstrate more stable training and improved performance on both in-distribution, GSM8K~\cite{GSM8K} and MATH~\cite{MATH}, and out-of-distribution benchmarks, AIME 2024~\cite{aime2024}, AIME 2025~\cite{aime2025} and AMC 2023~\cite{amc2023}.
    In addition, ablation studies confirm the importance of each term of GTPO.

\end{itemize}
The paper is structured as follows: Section~\ref{sec:related} reviews related work, and Section~\ref{sec:grpo_loss} introduces key preliminaries. Section~\ref{sec:grpo_problems} analyzes GRPO’s limitations, while Section~\ref{sec:methodology} presents our GTPO approach. Section~\ref{sec:experiments} reports experimental results, and Section~\ref{sec:conclusion} concludes the paper.

\section{Related Work}\label{sec:related}
\normalsize
\noindent \textbf{Reinforcement Learning in LLMs.} RL has been widely adopted in decision-making tasks \cite{google-rl,rlhf-notLLM,dota-2}, and nowadays is increasingly applied to the alignment and fine-tuning of LLMs. A prominent approach is RL from Human Feedback (RLHF), introduced in InstructGPT \cite{rlhf}, and further developed by Anthropic \cite{antropic-rlhf}. RLHF has become central to the training pipelines of state-of-the-art LLMs such as Claude 3 \cite{anthropic2024claude3}, Gemini \cite{gemini}, and GPT-4 \cite{GPT4}. It typically includes supervised fine-tuning, a reward model, and the adoption of Proximal Policy Optimization (PPO) \cite{ppo}. Here, PPO improves training stability by constraining updates through a clipped surrogate objective, offering a more practical alternative to Trust Region Policy Optimization (TRPO) \cite{TRPO}. However, it remains sensitive to reward scaling and can suffer from training instability \cite{trgppo,PPOissues2,PPOissues}, requiring multiple refinements \cite{ppo37implementations}. Therefore, multiple variations have been proposed over the years, such as TRGPPO \cite{trgppo}, alphaPPO \cite{alphaPPO}, and PPO-ALR \cite{PPO-ALR}.

\textbf{Advancements and Limitations in GRPO.} To overcome the need for a critic model, Deepseek introduced GRPO \cite{GRPO,GRPO-r1}, an approach that, given a question, compares multiple responses (completions) to derive relative rewards. GRPO demonstrates state-of-the-art performance on math benchmarks and achieves human-like alignment without relying on explicit manual feedback or critic networks \cite{grpo-alignment-human}.
Despite its promise, potential limitations of GRPO have been recently emerged, as bias effects \cite{overfit-grpo}, gradient imbalance \cite{low-token-grpo-issue}, which lead to undertraining of rare yet informative tokens \cite{liu2025understanding},
and degradations (or even collapse) of model performance like in PPO~\cite{dohare2023overcoming}. 
Recently, Group Sequence Policy Optimization (GSPO)~\cite{zheng2025group} has been proposed as a GRPO variant that performs sequence-level optimization to improve stability and efficiency, while coinciding with GRPO at the first iteration.

We identify two key issues in GRPO: token-level penalization and policy collapse, for which we show that KL divergence provides limited protection, while entropy-based analysis offers clearer corrective signals~\cite{cui2025entropymechanismreinforcementlearning}. These insights motivate GTPO, which improves stability and performance during training and evaluation.

\section{Preliminaries}\label{sec:background}

\label{sec:grpo_loss}

During training, given an input prompt $q$, the sampling LLM 
$\pi_{\theta_{\text{old}}}(\cdot \mid q)$ generates a group of $G$ 
completions $\{o_i\}_{i=1}^G$, where each completion consists of tokens 
denoted by $o_{i,t}$. Each completion receives a scalar reward $R_i$, combining accuracy and 
formatting verifiable scores. These rewards are then normalized into advantages $\hat{A}_i = (R_i - \bar{R}) / {\mathrm{std}(R)}$.
where $\bar{R}$ and $\mathrm{std}(R)$ denote the mean and standard deviation 
of the $R_i$ values within the group $G$. The GRPO objective is:
\footnotesize
\begin{equation}
\mathcal{J}_{\text{GRPO}}(\theta)
=
\mathbb{E}_{q,\{o_i\}} \left[
\frac{1}{G} \sum_{i=1}^{G} \bar{\mathcal{C}}_i
- \beta \cdot D_{\mathrm{KL}}(\pi_\theta \| \pi_{\mathrm{ref}})
\right],
\label{eq:grpo-full}
\end{equation}

\normalsize

\noindent where $D_{\mathrm{KL}}(\pi_\theta \| \pi_{\mathrm{ref}})$ refers to KL divergence scaled by $\beta$ and $\bar{\mathcal{C}}_i$  is the average over tokens:



\footnotesize
\begin{equation}
\bar{\mathcal{C}}_i = \sum_{t=1}^{|o_i|} \frac{\min\Big(
r_{i,t}(\theta)\,\hat{A}_i,\;
\operatorname{clip}\big(r_{i,t}(\theta), 1-\varepsilon, 1+\varepsilon\big)\,\hat{A}_i
\Big)}{|o_i|},
\label{eq:combined_objective}
\end{equation}

\normalsize
\noindent where the importance sampling ratio is defined as:
$
r_{i,t}(\theta) = \frac{\pi_\theta(o_{i,t} \mid s_{i,t})}{\pi_{\theta_{\text{old}}}(o_{i,t} \mid s_{i,t})},
$
\noindent and the state at step $t$ is $s_{i,t} = (q, o_{i,<t})$, with $o_{i,<t}$ denoting the sequence of tokens generated before step $t$ for completion $i$.

The policy $\pi_\theta$ is obtained from logits $f_\theta$ via
$\pi_\theta(o_{i,t}\mid s_{i,t}) = \mathrm{softmax}(f_\theta) \in \mathbb{R}^V$,
where $V$ is the vocabulary size. The KL term in Eq.~\eqref{eq:grpo-full} penalizes deviations from a reference policy $\pi_{\mathrm{ref}}$, scaled by $\beta$. For clarity, in the gradient expression we omit the clipping and keep only the dependence on the ratio $r_{i,t}$ (full derivation in App. \ref{appendix:aggregated-grpo-loss}). Under the standard policy-gradient approximation,

\small
\begin{align}
\nabla_\theta \mathcal{J}_{\text{GRPO}}(\theta)
& = \mathbb{E}_{q,\{o_i\}} \Bigg[\frac{1}{G} \sum_{i=1}^{G} \Bigg( \frac{\hat{A}_i}{|o_i|} \sum_{t=1}^{|o_i|}
 g_{i,t} \notag \\&
\;-\; 
\beta \cdot \nabla_\theta \mathrm{D}_{\mathrm{KL}}(\pi_\theta \| \pi_{\mathrm{ref}})\Bigg)\Bigg]
\label{eq:grpo_complete_gradient}
\end{align}

\normalsize
Where the token-level contribution $g_{i,t}$ depends only on the ratio and on the policy gradient:
\begin{equation}
g_{i,t}
:=
r_{i,t}(\theta)\,\nabla_\theta \log \pi_\theta(o_{i,t} \mid s_{i,t})
\end{equation}
Writing $g_{i,t}$ in terms of logits, let $j$ be the index of $o_{i,t}$, with probability $\pi_\theta^j$; then
\small
\begin{equation}
g_{i,t}
=
r_{i,t}(\theta)
\left[
\nabla_\theta f_\theta^j (1-\pi_\theta^j)
-
\sum_{k \ne j} \pi_\theta^k \,\nabla_\theta f_\theta^k
\right]
\end{equation}
\normalsize
A positive normalized advantage $\hat{A}_i > 0$ increases the logit of the selected token (scaled by $r_{i,t}$) while decreasing those of its alternatives; negative advantages have the opposite effect.

\section{GRPO Issues}\label{sec:grpo_problems}
We highlight two major limitations of GRPO: \textit{token-level penalization} and \textit{policy collapse}.

\subsection{Token-level Penalization}
\label{ss:token-level}

In this section we show that GRPO tends to have negative effects on the updates of tokens shared across multiple completions in the
same group even though they can be essential for reward maximization.


\noindent \textbf{Prefix Tokens Penalization.}
Given a prompt \( q \), the model generates a set of completions
$\{o_i\}_{i=1}^G$, which may begin with the same sequence of tokens and then diverge at a specific point, a token that acts as a crossroads. This behavior is typical in instruction-tuned models, where early tokens tend to be similar across completions~\cite{first-token-similar}. 
However, not all completions in the group \( G \) necessarily share the same prefix. In practice, we often observe multiple distinct prefixes, each shared by a subset of the completions.
To model this, we partition the \( G \) completions into \( K \) disjoint groups \(\mathcal{G}_1, \ldots, \mathcal{G}_K\) so that all completions in a group \(\mathcal{G}_k\) have a common prefix \( S_{\text{pfx}}^{(k)} \). If all \( G \) completions share the same prefix, then \( K = 1 \). 
Tokens in \( S_{\text{pfx}}^{(k)} \) are affected by the combined advantages of all completions in the group, while other tokens only depend on the advantage of their own completions. Under this structure, we rewrite the GRPO gradient (Eq.~\ref{eq:grpo_complete_gradient}) by summing the contributions of all groups, where each group contributes a prefix term and a set of individual terms. 
The gradient, excluding the KL term, becomes:


\small
\begin{equation}
\begin{split}
\nabla_\theta \mathcal{J}_{\text{GRPO}}(\theta)  & = \mathbb{E}_{q,\{o_i\}} \Bigg[\frac{1}{G} \sum_{k=1}^{K} \Bigg[
\sum_{i \in \mathcal{G}_k} \Bigg(
   \frac{\hat{\mathcal{A}}_i}{|{o}_i|}
   \sum_{j \notin S_{\text{pfx}}^{(k)}}
   g_{i,j}\Bigg) 
   \\ & + \smash{\underbrace{
       \sum_{i \in \mathcal{G}_k}\Bigg(
       \frac{\hat{\mathcal{A}}_i}{|{o}_i|}
       \sum_{t \in S_{\text{pfx}}^{(k)}}
       g_{i,t} }_{\Delta_{\text{pfx}}^{(k)}}}\Bigg) 
\quad
\Bigg] \Bigg]
\end{split}
\end{equation}

\normalsize
The term \( \Delta_{\text{pfx}}^{(k)} \) represents the gradient update associated with the shared prefix of group \( \mathcal{G}_k \). If a completion does not share any prefix with others, then \( \mathcal{G}_k \) contains only that single completion, resulting in \( S_{\text{pfx}}^{(k)} = \emptyset \) and \( \Delta_{\text{pfx}}^{(k)} = 0 \). Since the log-probability terms \( g_{i,t} \) are identical and so constant across completions in the group \( S_{\text{pfx}}^{(k)} \), the gradient component \( \Delta_{\text{pfx}} \) is primarily driven by the sum of normalized advantages \( \sum_{i \in \mathcal{G}_k} \hat{\mathcal{A}}_i / |{o}_i| \). This implies that the update to a prefix mainly depends on the relative balance of advantages across all completions in group \( k \). 
For instance, the sum can be negative, when correct completions (with positive advantages) are generally longer than incorrect ones. In such cases, the denominator \( |{o}_i| \) for the correct completions becomes larger, reducing their contribution to the overall gradient. As a result, the model may be penalized for generating desirable and beneficial prefix tokens.
This introduces a bias against shared initial tokens, such as formatting tokens or reasoning tags (\texttt{<reasoning>}), essential for the response structure and correctness.
\textbf{Dependencies From Completion Advantage.}
GRPO can also induce undesirable penalization on tokens that appear after the prefix.  
Certain tokens, in particular formatting ones, may occur in multiple completions at varying positions and within different contexts, some correct, others incorrect, making them more susceptible to inconsistent and potentially harmful updates. This induces potential issues in the update. For instance, if a token \( \tau \) frequently appears in completions with negative advantage, its probability may be reduced, even if the token is syntactically correct and required.
Additionally, further issues can arise when a completion only partially follows the expected format. For example, if a completion closes a first part correctly with a formatting token $\tau$ (e.g. \texttt{</reasoning>}) but then omits \texttt{<answer>}, $\tau$  may receive a low or even negative reward. This, in turn, penalizes \texttt{</reasoning>}, despite it being a correct token in the completion. This occurs because the update for each token primarily depends on the advantage assigned to the entire completions in which it appears, without directly considering whether the token's individual contribution was beneficial. This phenomenon is particularly impactful in the case of formatting tokens, as illustrated in the previous examples, and in the case of shared final tokens that form a common suffix among completions.

\subsection{Policy Collapse} \label{ss:policy_collapse}
The GRPO gradient, Eq.~\eqref{eq:grpo_complete_gradient}, highlights a dependency on both the advantage and the probability scores assigned by the policy \( \pi_\theta \).  
To gain a deeper understanding of how the advantage influences the learning dynamics, we analyze the average Shannon entropy of the output distribution \( \pi_\theta \) over completions \( i \) and tokens \( t \) during training. The Shannon entropy is defined as  
\(
H(\mathbf{p}) = -\sum_{j=1}^{n} p_j \ln p_j,
\)
where \( \mathbf{p} \) is a probability vector. For a token position $t$ in a completion \( o_i \), we consider \( \mathbf{p} = \pi_\theta(o_{i,t}|s_{i,t}) \), and we define \( \langle H \rangle_i \) as the average entropy across in the completion \( o_i \).
At a generation step, let us consider a low entropy (e.g., \( \langle H \rangle_i  \ll \ln 2 \))
\footnote{An average entropy close to \(\ln 2\) indicates that the model is choosing between two equiprobable tokens at each step.
}
, indicating that the model is highly confident, i.e., it assigns almost all probability mass to a single token index $j$, with \( \pi_\theta^j \approx 1 \) and \( \pi_\theta^k \approx \epsilon \ll 1 \), where $k \neq j$. If this token leads to a negative outcome (\( \hat{\mathcal{A}}_i < 0 \)), GRPO penalizes it sharply, since the gradient of the GRPO loss at index \( \tau \) for completion \( i \) becomes:

\footnotesize
\begin{equation}
\nabla_\theta {\mathcal{J}}^{(i,\tau)}_{\text{GRPO}}(\theta) \approx
\begin{cases} 
- \dfrac{|\hat{\mathcal{A}}_i|}{|o_i|} r_{j,i}(\theta) \nabla_\theta f_\theta^j \cdot (1 - \epsilon) &\text{for } j \\
\;\;\dfrac{|\hat{\mathcal{A}}_i|}{|o_i|}  r_{k,i}(\theta) \nabla_\theta f_\theta^k \cdot \epsilon &\text{for } k \ne j
\end{cases}
\label{eq:collapse}
\end{equation}

\begin{figure}
    \centering
\includegraphics[width=\linewidth]{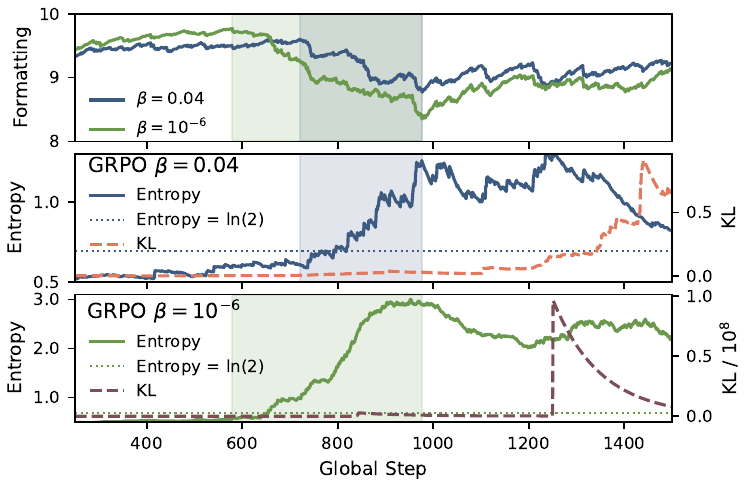}
    \vspace{-1em}
    \caption{\small{GRPO training of LLaMA 8B on GSM8K \( G = 8 \) comparing \( \beta = 0.04 \) and \( \beta = 10^{-6} \). Plots show formatting reward (top), average entropy (middle), and KL divergence (bottom).}}
    \label{fig:kl-entropy}
\end{figure}

\normalsize
Note that the KL term has been omitted here, its contribution will be addressed next. Both $r_{k,i}(\theta)$ and $r_{j,i}(\theta)$ are equal to $1$ at the first iteration and remain approximately $1$ in the subsequent ones. This results in a negative update for the selected token \( j \), and small positive updates for all other tokens \( k \ne j \), even if they are potentially syntactically or semantically incorrect. While each of these positive updates is individually small, they can accumulate over time, gradually increasing the probability of initially implausible tokens.
The effect is further amplified by \textit{GRPO’s zero-mean constraint} (the average of the advantages of the group $\bar{A}=0$): when most completions receive positive rewards, the few with negative rewards must carry disproportionately large negative advantages to maintain balance. These penalties can suppress correct predictions and unintentionally amplify the likelihood of undesirable alternatives, raising entropy and introducing instability.
Over time, this harms output quality and increases the risk of introducing structural and semantic errors that propagate through subsequent training steps. As the model drifts away from desirable behaviors, the reward signal becomes less meaningful. This phenomenon is illustrated in the top plot of Figure~\ref{fig:kl-entropy}, which shows the average formatting reward during the training of LLaMA 8B on GSM8K, using GRPO with \(\beta = 0.04\) (as originally proposed in~\cite{GRPO}) and \(\beta = 10^{-6}\). At the beginning of training, the formatting reward increases more rapidly with \(\beta = 10^{-6}\), while it remains relatively flat with \(\beta = 0.04\). However, in both cases, the reward begins to drop below 9, after 580 steps for \(\beta = 10^{-6}\) and around step 750 for \(\beta = 0.04\).

\noindent \textbf{KL Reacts Late to Avoid Collapse.} 
Middle and bottom plots of Figure~\ref{fig:kl-entropy} show the average entropy \( \langle H \rangle_i \) and average KL divergence, computed on the generated completions, for the two previous training runs of LLaMA 8B. Both KL divergence curves remain stable up to approximately step 1200, well after the degradation points, when the formatting rewards have already collapsed and entropy levels are high. 
Only around the KL peak, the reward curves show signs of stabilization. 
In contrast, the entropy curves begin to rise sharply as the formatting rewards start to decline, and continue to increase as the rewards deteriorate further.
This suggests that the KL divergence acts as a delayed corrective signal, rising only after collapse. In contrast, entropy tracks the policy collapse in real time, increasing as the policy loses structure. 


\section{Method}
\label{sec:methodology}
This section presents GTPO, which mitigates gradient conflicts and prevents policy collapse.

\subsection{Conflict-Aware Gradient Correction}
As discussed in Section~\ref{ss:token-level}, The model can encounter gradient conflict issues, where shared tokens receive both positive and negative updates depending on the advantage assigned to each completion.
Thus, we propose a methodology aimed at selectively masking gradient contributions for tokens involved in conflicting updates.
To this end, we first identify \textit{conflict tokens}, as tokens that appear in the same position, either from the beginning or the end, across completions with both positive and negative advantages. As discussed in the context of GRPO's issues part, these tokens often receive conflicting gradient updates during training. Then, we mitigate such conflicts by correcting their gradient updates accordingly.


\noindent \textbf{Conflict tokens definitions.}
Let $\{o_i\}_{i=1}^G$ be a group of completions. Each completion \(o_i\) contains a sequence of tokens $\{o_{i,t}\}_{t=1}^{|o_i|}$, and is associated with an advantage value \(A^{(i)} \in \mathbb{R}\). We define $G^-$ and $G^+$ as the sets of completions with $\mathcal{A} < 0$ and $\mathcal{A} > 0$, respectively.

\noindent \textit{Left-to-right alignment.}  
A token \(v \in \mathcal{V}\) is a \emph{forward conflict token} at position \(p\) if it appears at the same position in at least one completion with positive advantage and in at least one with negative:
\[
\exists\, i \in G^+: \ o_{i,p} = v
\quad \wedge \quad
\exists\, j \in G^-: \ o_{j,p} = v.
\]

\noindent \textit{Right-to-left alignment.} A token \( v \) is defined as a \emph{backward conflict token} at offset \( r \) if it occurs at the \( r \)-th position from the end in at least one completion with a positive advantage and in at least one with a negative advantage:

\footnotesize
\[
\exists\, i \in G^+:\ o_{i,|o_i| - r} = v
\quad \wedge \quad
\exists\, j \in G^-:\ o_{j,|o_j| - r} = v.
\]

\normalsize
\noindent \textbf{Gradient Reweighting with conflict masks.}
We construct binary masks for tokens across completions to target positions potentially affected by gradient conflicts and thus correct their updates.
We first define a \textit{forward mask} \( \mathcal{M}^{\mathrm{fw}}_i \in \{0,1\}^{|o_i|} \) by scanning \( o_i \) from left to right, setting 1 over the first contiguous span of forward conflict tokens and 0 elsewhere.  
The \textit{backward mask} \( \mathcal{M}^{\mathrm{bw}}_i \) is obtained by scanning from right to left, marking the first contiguous span of backward conflict tokens.  
A \textit{final mask} $\mathcal{M}_i$ is then defined as:
\(
\mathcal{M}_i = \mathcal{M}^{\mathrm{fw}}_i \lor \mathcal{M}^{\mathrm{bw}}_i,
\)
highlighting only the initial and final conflict regions in each completion $o_i$. 
After computing each \( \mathcal{M}_i \), we correct inconsistent gradient updates over the selected conflict tokens, while leaving other token updates unchanged, thought the following preliminary token-level loss:

\begin{align}
\mathcal{J}^* = \mathbb{E}_{q,\{o_i\}} \left[\frac{1}{G} \sum_{i=1}^{G} \frac{\mathcal{A}_i}{|o_i|}
\sum_{t = 1}^{|o_i|} r_{i,t} \lambda_{i,t} \right]
\label{eq:first_formula}
\end{align}

\noindent where \(\lambda_{i,t}\) controls the update of each token based on its conflict position and the sign of the advantage $\mathcal{A}_i$:
\begin{equation}
\lambda_{i,t} =
\begin{cases}
    1 & \text{if } \mathcal{M}_{i,t} = 0,\\
    0 & \text{if } \mathcal{M}_{i,t} = 1 \text{ and } \mathcal{A}_i < 0, \\
    2 & \text{if } \mathcal{M}_{i,t} = 1 \text{ and } \mathcal{A}_i > 0.
\end{cases}
\label{eq:lambda}
\end{equation}
Intuitively, the mask disables negative gradients on conflict tokens, and instead reinforces them only if they appear in positively rewarded completions. Note that the total signal magnitude is preserved across the group: since \(\sum_{i \in G^+} |\mathcal{A}_i| = \sum_{i \in G^-} |\mathcal{A}_i|\), doubling the signal for positive completions compensates for the removal of negative updates, maintaining training stability while preventing semantic drift. 
Additional details and derivations of the weighting scheme are provided in appendix \ref{sec:full-der-gtpo}, while its benefits are evaluated in the ablation studies (Section~\ref{sec:ablation}). Importantly, the final mask \(  \mathcal{M}_i \) targets only the initial and final contiguous conflict tokens to preserve the semantic structure of completions. In fact, masking isolated conflict tokens in the middle could harm stability and learning, as their meaning often depends on surrounding context, and altering their gradients may be counterproductive. In contrast, the outer spans typically correspond to formatting tags, such as \texttt{<reasoning>} or \texttt{</answer>}, which are the primary source of conflict in GRPO, as discussed in Section \ref{ss:token-level}. Focusing the correction on these regions protects structural tokens without interfering with the central part of the completion, where meaningful differences in trajectories emerge.


\subsection{Entropy Control}
As discussed in Section \ref{ss:policy_collapse}, GRPO can lead to policy collapse, where standard KL term may react too slowly. To address this, we propose entropy-based regularization terms during training. These consist of two key parts: (i) a filtering mechanism to discard unstable completions, and (ii) a regularization term that penalizes high-entropy behavior.

\noindent \textbf{Completion filter.}
Based on the policy-collapse analysis, we observe that high-entropy completions can jeopardize training by signaling structural uncertainty, particularly in models that naturally exhibit low average entropy. Applying gradients in such cases risks amplifying uncertainty and accelerating collapse. To mitigate this, we propose filtering out high-entropy completions, focusing on models prone on collapse against this. 
We define \( \langle H \rangle_{\text{ini}} \) as the model's initial entropy over a set of questions, measured prior to training. If \( \langle H \rangle_{\text{ini}} < \ln 2 \), we assume that the model tends to produce low-entropy outputs, making it more sensitive to high-entropy completions during training. In this case, we apply an entropy-based filtering mask \( \delta_i \) that filter out the associated advantage signal. The mask $\delta_i$ is formally defined as:

\small
\begin{equation}
  \delta_i =
  \begin{cases}
    1, & \text{if } \langle H \rangle_{ini} > \ln 2, \\
    0, & \text{if } \langle H \rangle_{ini} < \ln 2 \text{ and } \langle H \rangle_i > \ln 2, \\
    1, & \text{if } \langle H \rangle_{ini} < \ln 2 \text{ and } \langle H \rangle_i \leq \ln 2.
  \end{cases}
\label{eq:filtering}
\end{equation}

\normalsize
%

\noindent \textbf{Entropy Regularization.}
Inspired by PPO~\cite{entropy-on-ppo,ppo37implementations}, we add a regularization term based on the average tokens entropy of each completion, \( \langle H \rangle_i \), where \( \gamma \) balance the importance of this term in the final loss, as shown below. Note that, based on the internal characteristics of GRPO to implicitly increase entropy over time, we decided to minimize the term. This acts as a way to reduce the model entropy over time. The combination of these entropy-based strategies and the token-level loss defines the final GTPO objective, as:

\footnotesize
\begin{align}
\mathcal{J}_{\text{GTPO}} = \mathbb{E}_{q,\{o_i\}} \left[\frac{1}{G} \sum_{i=1}^{G} \frac{\delta_i \cdot (\mathcal{A}_i - \gamma \cdot \langle H \rangle_i)}{|o_i|}
\sum_{t = 1}^{|o_i|} r_{i,t}\lambda_{i,t} \right]
\label{eq:final-our}
\end{align}

\normalsize
In contrast to vanilla-GRPO~\cite{GRPO}, shown in Equation~\ref{eq:grpo-full}, the proposed GTPO Objective does not
 rely on KL divergence, so the reference model
 is not required during training, making the process more lightweight and faster. Others~\cite{hu2025open, liu2025understanding} tried to remove the KL
term from GRPO objective, but our tests (see Section~\ref{sec:experiments}) show it
 is sometimes necessary to get a more reliable and
 stable GRPO training.
The benefits of each component, and hyperparameter of GTPO Objective are discussed and demonstrated, also through ablation studies in the following experimental section.

\begin{figure*}[t]
  \centering
\includegraphics[width=\textwidth]{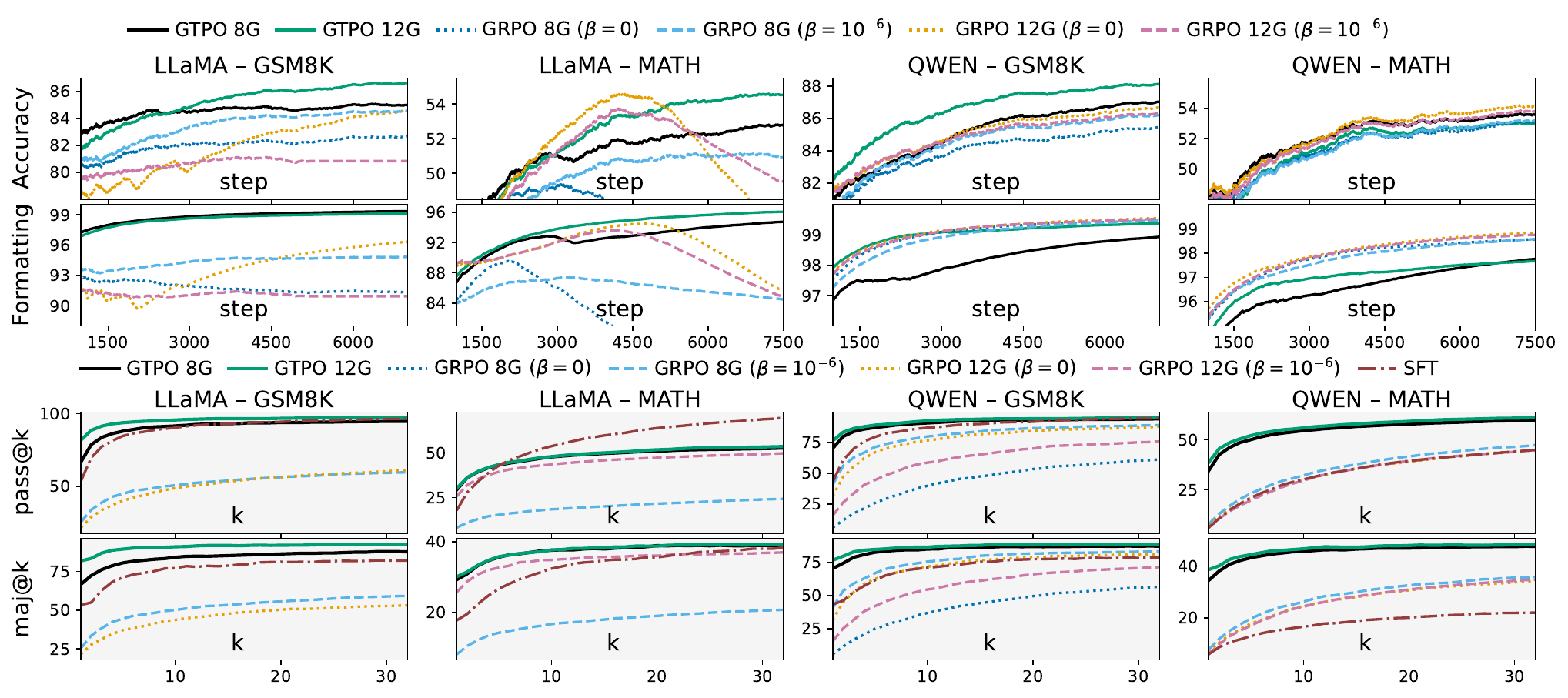}
  \caption{\small{\textit{Training accuracy and formatting rewards} (\%) of GTPO and GRPO over training steps on MATH and GSM8K (top).\textbf{ In-distribution evaluation} of models trained with GTPO, GRPO, and SFT on the corresponding test sets, using \texttt{pass@k} and \texttt{maj@k} (\%) over \( k \) (bottom).}}
  \label{fig:tuttapagina}
\end{figure*}

\section{Experiments}
\label{sec:experiments}

\begin{figure*}[t]

    \begin{subfigure}[b]{\textwidth}
    \centering
    \includegraphics[width=0.9\textwidth]{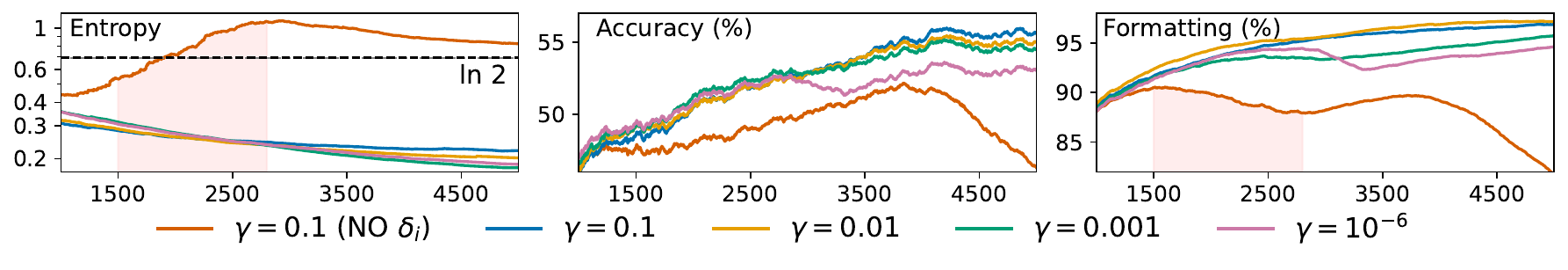}
    \caption{\footnotesize{LLAMA GTPO train on MATH with different \(\gamma\): entropy, accuracy, and formatting. ``NO \(\delta_i\)'' disables entropy filtering.}}
    \label{fig:entropy_regularization}
    \end{subfigure}

        \vspace{-0.05cm}

    \centering
    \begin{subfigure}[b]{\textwidth}
        \centering
        \includegraphics[width=0.9\textwidth]{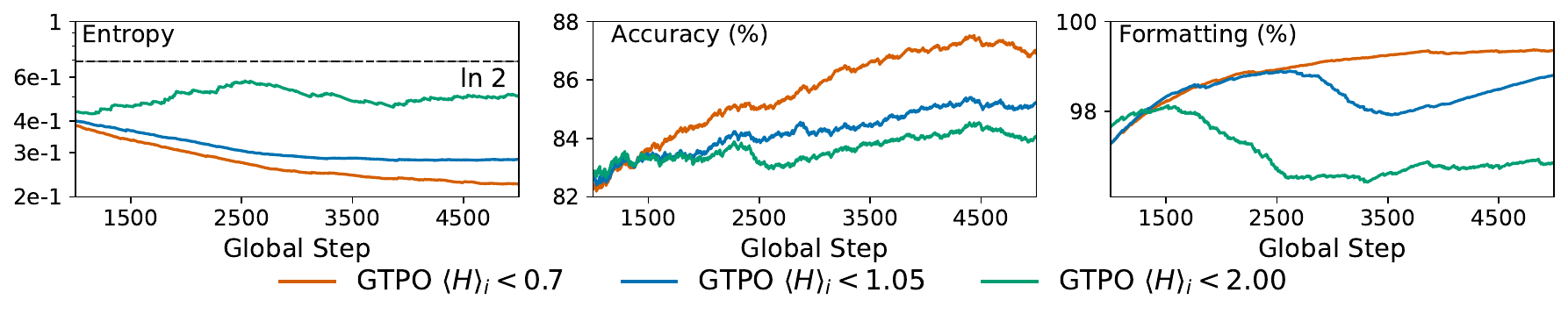}

        \caption{\footnotesize{Entropy, accuracy, and formatting in LLaMA trained with GTPO on GSM8K with different values of $\langle H \rangle_{i}$.}}
        \label{fig:ablation_ini_llama}
    \end{subfigure}

    \vspace{-0.05cm}

    \begin{subfigure}[b]{\textwidth}
        \centering
        \includegraphics[width=0.9\textwidth]{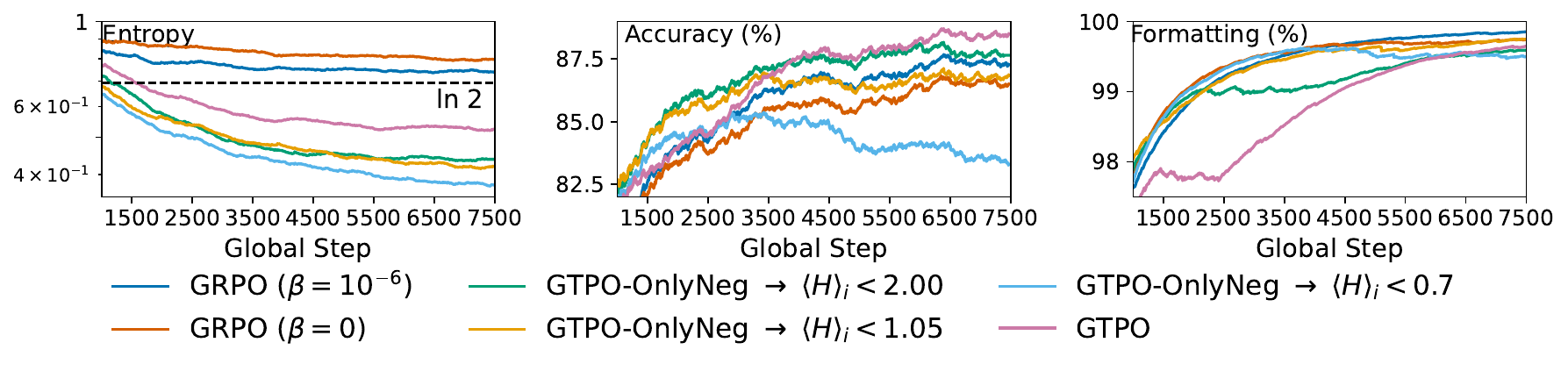}

        \caption{\footnotesize{QWEN on GSM8K: GRPO/GTPO with different $\langle H \rangle_{i}$ thresholds.  \textit{OnlyNeg}} means $\langle H \rangle_{i}$ only for completions with $\mathcal{A}^-$.} 
        \label{fig:ablation_only_neg}
    \end{subfigure}

    \vspace{-0.05cm}

    \begin{subfigure}[b]{\textwidth}
        \centering
        \includegraphics[width=0.9\textwidth]{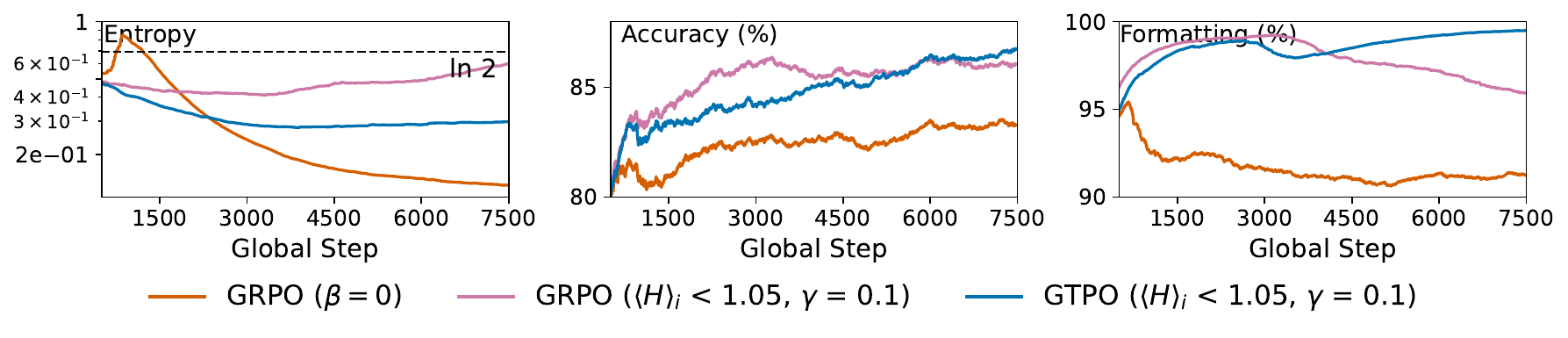}

        \caption{\footnotesize{LLaMA trained on GSM8K with GRPO, and with GRPO/GTPO under the constraint $\langle H \rangle_{i} < 1.05$ and \(\delta_i=0.1\).}}
        \label{fig:ablation_grpo_our}
    \end{subfigure}

    \caption{\small{Comparison of  entropy, accuracy, formatting under different training settings for GTPO and GRPO.}}
\end{figure*}

We conducted our experiments using LLaMA 8B and Qwen 2.5 (3B). After training the models on the GSM8K and MATH training sets, we evaluated them on the corresponding test splits (in-distribution evaluation). We further assessed performance using the AIME 2024, AIME 2025, and AMC 2023 benchmarks (out-distribution evaluation). For comparison, all models were also trained using SFT and GRPO (with both \( \beta = 0 \) and \( \beta = 10^{-6} \) to assess the impact of the KL term)\footnote{\( \beta = 10^{-6} \) yielded the best preliminary results.}. Both GRPO and GTPO were evaluated with two generation sizes: \( G = 8 \) and \( G = 12 \).
For GTPO, we compute \( \langle H \rangle_{\text{ini}} \) by evaluating the original model entropy on the first 100 samples of the training set and we chose 0.1 for $\gamma$ to reduce entropy suppression (see Figure~\ref{fig:entropy_regularization}). Due to computational cost and training time, one iteration is adopted in Eq.~\ref{eq:grpo-full}~\cite{simoni2025improving}, where $\pi_\theta = \pi_{\theta_{\text{old}}}$, making the clipping unnecessary (GRPO objective becomes Eq.~\ref{eq:reward-term-grpo}). All training was performed with a learning rate of \( 10^{-6} \), with temperature set to 1.0. Other training hyperparameters are shown in Listing~\ref{lst:training-settings} (Appendix~\ref{sec:full-der-gtpo}). We conducted all the experiments on 2 NVIDIA A100. In Section~\ref{sec:exp_first}, we show the comparison between GTPO, GRPO, and SFT on both in-distribution and out-of-distribution benchmarks. In Section~\ref{sec:ablation}, we provide ablation studies that isolate the contribution of each component of the GTPO objective.

\begin{table*}[t] 
\centering
\resizebox{\textwidth}{!}{%
    \begin{tabular}{lll ccccc J ccccc J ccccc J ccccc}
    \toprule
    & & & \multicolumn{5}{c}{\textbf{AIME 2024}} & & \multicolumn{5}{c}{\textbf{AIME 2025}} & & \multicolumn{10}{c}{\textbf{AMC 2023}} \\
    \cmidrule(lr){4-8} \cmidrule(lr){10-14} \cmidrule(lr){16-26}
    
    & & & \multicolumn{5}{c}{Pass@k (\%)} & & \multicolumn{5}{c}{Pass@k (\%)} & & \multicolumn{5}{c}{Pass@k (\%)} & & \multicolumn{5}{c}{Maj@k (\%)} \\
    \cmidrule(lr){4-8} \cmidrule(lr){10-14} \cmidrule(lr){16-20} \cmidrule(lr){22-26}
    
    \textbf{Model} & \textbf{Dataset} & \textbf{Method} 
    & 1 & 8 & 16 & 32 & 64 &
    & 1 & 8 & 16 & 32 & 64 &
    & 1 & 8 & 16 & 32 & 64 &
    & 4 & 8 & 16 & 32 & 64 \\
    \midrule

    \multirow{6}{*}{\textbf{\rotatebox{45}{LLAMA}}} 
    & \multirow{3}{*}{\textbf{GSM8K}} 
      & SFT   & 0.0 & 0.0 & 0.0 & 13.3 & 20.0 & & 0.0 & 0.0 & 0.0 & 0.0 & 6.7 & & 12.5 & 37.5 & 42.5 & 57.5 & 72.5 & & 10.0 & 15.0 & 15.0 & 20.0 & 17.5 \\
    & &  GRPO  & 0.0 & 0.0 & 0.0 & 0.0 & 0.0 & & 0.0 & 0.0 & 0.0 & 00 & 0.0 & & 2.5 & 2.5 & 5.0 & 5.0 & 5.0 & & 2.5 & 2.5 & 5.0 & 5.0 & 5.0 \\
    & &\cg \textbf{GTPO}  &\cg 0.0 &\cg \textbf{10.0} &\cg \textbf{13.3} &\cg \textbf{20.0} &\cg \textbf{30.0} &\cg &\cg 0.0 &\cg \textbf{6.7} &\cg \textbf{6.7} &\cg \textbf{6.7} &\cg \textbf{13.3} &\cg &\cg \textbf{17.5} &\cg \textbf{47.5} &\cg \textbf{55.0} &\cg \textbf{65.0} &\cg \textbf{80.0} &\cg &\cg \textbf{25.0} &\cg \textbf{27.5} &\cg \textbf{35.0} &\cg \textbf{37.5} &\cg \textbf{42.5} \\
    \cmidrule(l){2-26} 
    
    & \multirow{3}{*}{\textbf{MATH}}  
      & SFT   & 0.0 & 3.3 & 10.0 & 20.0 & 26.7 & & 0.0 & 0.0 & 3.3 & 6.7 & \textbf{16.7} & & 10.0 & 45.0 & 60.0 & 65.0 & 80.0 & & 12.5 & 15.0 & 12.5 & 15.0 & 17.5 \\
    & & GRPO  & 0.0 & \textbf{13.3} & \textbf{16.7} & 20.0 & 36.7 & & 0.0 & 0.0 & 0.0 & 3.3 & 6.7 & & 20.0 & 37.5 & 40.0 & 52.5 & 70.0 & & 25.0 & 30.0 & 30.0 & 30.0 & 32.5 \\
    & &\cg \textbf{GTPO}  &\cg \textbf{3.3} &\cg \textbf{13.3} &\cg \textbf{16.7} &\cg \textbf{36.7} &\cg \textbf{43.3} &\cg &\cg 0.0 &\cg 0.0 &\cg \textbf{10.0} &\cg \textbf{10.0} &\cg 13.3 &\cg &\cg \textbf{22.5} &\cg \textbf{52.5} &\cg \textbf{67.5} &\cg \textbf{75.0} &\cg \textbf{82.5} &\cg &\cg \textbf{37.5} &\cg \textbf{37.5} &\cg \textbf{37.5} &\cg \textbf{37.5} &\cg \textbf{37.5} \\
    \midrule

    \multirow{6}{*}{\textbf{\rotatebox{45}{QWEN}}} 
    & \multirow{3}{*}{\textbf{GSM8K}} 
      & SFT   & 0.0 & 0.0 & 0.0 & 0.0 & 6.7 & & 0.0 & 0.0 & 0.0 & 6.7 & 6.7 & & 0.0 & 10.0 & 17.5 & 52.5 & 62.5 & & 0.0 & 2.5 & 7.5 & 7.5 & 7.5 \\
    & & GRPO  & 0.0 & 3.3 & 6.7 & \textbf{16.7} & 20.0 & & 0.0 & 3.3 & 6.7 & 10.0 & 20.0 & & 7.5 & 40.0 & 52.5 & 67.5 & 77.5 & & 17.5 & 12.5 & 25.0 & 30.0 & 42.5 \\
    & &\cg \textbf{GTPO}  &\cg \textbf{3.0} &\cg \textbf{13.3} &\cg \textbf{16.7} &\cg \textbf{16.7} &\cg \textbf{26.7} &\cg &\cg 0.0 &\cg \textbf{6.7} &\cg \textbf{10.0} &\cg \textbf{20.0} &\cg \textbf{26.7} &\cg &\cg \textbf{37.5} &\cg \textbf{65.0} &\cg \textbf{70.0} &\cg \textbf{85.0} &\cg \textbf{95.0} &\cg &\cg \textbf{40.0} &\cg \textbf{42.5} &\cg \textbf{47.5} &\cg \textbf{50.0} &\cg \textbf{45.0} \\
    \cmidrule(l){2-26}
    
    & \multirow{3}{*}{\textbf{MATH}}  
      & SFT   & 0.0 & 0.0 & 0.0 & 0.0 & 6.7 & & 0.0 & 0.0 & 0.0 & 0.0 & 3.3 & & 0.0 & 12.5 & 35.0 & 47.5 & 65.0 & & 0.0 & 2.5 & 7.5 & 7.5 & 7.5 \\
    & & GRPO  & 0.0 & 3.3 & 3.3 & 6.7 & 16.7 & & 0.0 & 3.3 & \textbf{10.0} & \textbf{13.3} & 16.7 & & 30.0 & 67.5 & 70.0 & 77.5 & \textbf{87.5} & & 27.5 & 42.5 & \textbf{42.5} & 45.0 & 42.5 \\
    & & \cg \textbf{GTPO}  & \cg \textbf{3.3} & \cg \textbf{6.7} & \cg \textbf{13.3} & \cg \textbf{20.0} & \cg \textbf{30.0} & \cg & \cg \textbf{3.3} & \cg \textbf{10.0} & \cg \textbf{10.0} & \cg \textbf{13.3} & \cg \textbf{23.3} & \cg & \cg \textbf{40.0} & \cg \textbf{75.0} & \cg \textbf{80.0} & \cg \textbf{82.5} & \cg 85.0 & \cg & \cg \textbf{52.5} & \cg \textbf{47.5} & \cg \textbf{42.5} & \cg \textbf{50.0} & \cg \textbf{50.0}\\
    \bottomrule
    \end{tabular}%
}

    \caption{\small{Qwen and LLaMA 8B \textbf{out-of-distribution evaluation} using the best of GTPO, GRPO, and SFT on AIME 2024/25 and AMC, reporting \texttt{pass@k} (and \texttt{maj@k} for AMC), with $1 \leq k \leq 64$.}}
\label{tab:ood_overall}
\end{table*}

\subsection{Performance Evaluation}
\label{sec:exp_first}
\noindent \textbf{Training Dynamics of GTPO and GRPO.}
In Figure~\ref{fig:tuttapagina}-top (the first two rows), we compare the training stability and performance of GTPO against GRPO. Formatting and accuracy rewards are reported as percentages, where the maximum value (100\%) corresponds to a reward of 10 in both. 
On the GSM8K dataset with LLaMA, GTPO consistently outperforms GRPO across all training steps in both accuracy and formatting metrics. On the more challenging MATH dataset, GRPO (with \( G = 12 \) and \(\beta=0\)) initially achieves slightly better accuracy around the midpoint of training. However, its performance drops sharply in the second half of training due to policy collapse, affecting both accuracy and formatting. 
In contrast, GTPO continues to improve steadily throughout training, avoiding collapse and maintaining stable performance.
For Qwen 2.5, on both GSM8K and MATH, GTPO achieves comparable or improved accuracy relative to GRPO, with only a slight decrease in formatting performance (still above 97\% in all runs). Note that, consistent with the analysis in Section~\ref{ss:policy_collapse}, GRPO training curves for Qwen 2.5 are not affected by a policy collapse, as it shows high-entropy behavior.
Overall, GTPO has more stable and reliable training compared to GRPO (see Fig.~\ref{fig:combined_math_qwen_gtpo} in Appendix~\ref{sec:complementary} for more details). 

\noindent \textbf{In-distribution Evaluation.} 
 The trained LLMs were evaluated on the test sets of GSM8K and MATH using the \texttt{pass@k}~\cite{pass@k} and \texttt{maj@k}~\cite{maj@k} metrics. The former measures whether at least one of the top-\( k \) completions yields a correct answer, while the latter assesses correctness via majority voting over the top-\( k \) completions. Note that, to ensure a fair comparison, we evaluated GRPO on LLaMA using the model checkpoint corresponding to the training step with the highest accuracy reward, rather than the one obtained after policy collapse. Few curves are omitted because, even at their best checkpoint, the corresponding models were too unreliable and achieved consistently low performance. Figure~\ref{fig:tuttapagina} bottom (the last two rows) shows that GTPO consistently outperforms GRPO in almost all settings for both \texttt{pass@k} and \texttt{maj@k}, as \( k \) varies from 1 to 32. This indicates that models trained with GTPO exhibit stronger self-consistency when answering questions (higher \texttt{maj@k}) and better coverage of the correct answer across multiple completions (higher \texttt{pass@k}).
We also include testing comparisons with SFT, where GTPO consistently outperforms SFT in \texttt{maj@k} across all models and datasets, and achieves higher average performance in \texttt{pass@k}. Specifically, SFT surpasses GTPO only in \texttt{pass@k} for \( k > 5 \) on MATH with LLaMA, but not in terms of correctness with \texttt{maj@k}. Interestingly, in GTPO, larger values of \(G\) always lead to better performance, which is not the case for GRPO; in most cases, the latter benefits from a value of \(\beta = 10^{-6}\) compared to \(\beta = 0\).


\noindent \textbf{Out-of-distribution Evaluation.}
We further evaluate the models on three out-of-distribution benchmarks: AIME~2024, AIME~2025, and AMC~2023. We report \texttt{pass@k} for $k \in \{1, 8, 16, 32, 64\}$, and for AMC also \texttt{maj@k}, replacing $k=1$ with $k=4$ since $\texttt{maj@1}$ is identical to $\texttt{pass@1}$. For each dataset, we select the GRPO and GTPO variants with the value of $G$ that achieved the highest \texttt{pass@k} on the in-distribution tests.

Overall, GTPO achieves the strongest generalization performance across both model families. For LLaMA, GTPO consistently outperforms SFT and matches or surpasses GRPO for almost all values of $k$ on AIME~2024, AIME~2025, and AMC~2023. The only exception is AIME~2025 at $k=64$ when trained on MATH, where SFT is slightly higher. For Qwen, GTPO obtains the best results on nearly all metrics and especially excels on AMC, where it achieves the highest \texttt{pass@k} and \texttt{maj@k}. GRPO occasionally approaches or slightly exceeds GTPO at very large $k$ (e.g., AMC \texttt{pass@64} with MATH training), but SFT remains the weakest method overall, showing competitive performance only at large $k$.

\begin{figure}[t]
    \centering
    \includegraphics[width=0.48\textwidth]
    {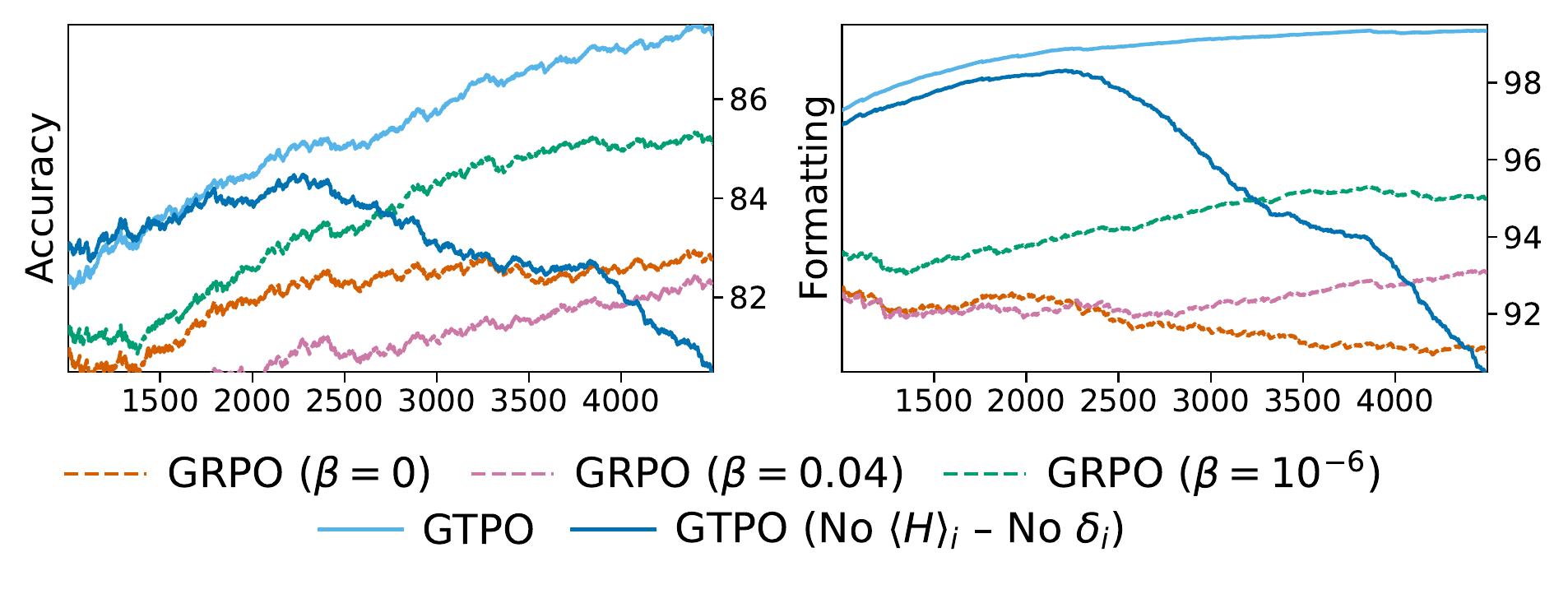}
    \vspace{-1em}
    \caption{\small{Accuracy and formatting of LLaMA on GSM8K. GRPO curves use different KL-\(\beta\) values, while \textit{``No \(\langle H \rangle_i\) – No \(\delta_i\)''} for GTPO applies only the conflict-aware gradient correction.
}}
\label{fig:total_vs_first_formula}
\end{figure}

\subsection{Ablation Studies}
\label{sec:ablation}

\normalsize
\noindent \textbf{Conflict-Aware gradient correction.} Figure~\ref{fig:total_vs_first_formula} shows the accuracy and formatting training curves of LLaMA on GSM8K. The model is trained using GRPO with KL \( \beta \) values set to 0, 0.04~\cite{GRPO}, and \( 10^{-6} \), while for GTPO, we use the full version  (Eq.~\ref{eq:final-our}) and a variant without entropy-based filtering and regularization (denoted as \textit{``No \(\langle H \rangle_i\) – No \(\delta_i\)''} in the figure). This latter configuration isolates the effect of GTPO when relying solely on the \textit{Conflict-Aware Gradient Correction} component (i.e., Eq.~\ref{eq:first_formula}).
As shown in the figure, GTPO outperforms GRPO in both accuracy and formatting during the first 2,500 steps. Beyond this point, GTPO with regularization and filtering continues to maintain better performance, whereas the variant without these components begins to degrade and eventually falls below GRPO. This behavior is expected, as the absence of regularization prevents the model from balancing the impact reward signals over time. Most importantly, before the policy collapse, the use of gradient correction alone yields higher rewards than GRPO.

\noindent \textbf{Impact of Entropy Regularization.}
Figure~\ref{fig:entropy_regularization} shows the training curves for average completion entropy (left), accuracy (middle), and formatting (right) on LLaMA~8B trained on MATH. We compare several entropy regularization strengths (\(\gamma = 0.1, 0.01, 0.001, 10^{-6}\)) and a variant with \(\gamma = 0.1\) but without entropy filtering (``\textit{NO}~\(\delta_i\)''). Higher values of \(\gamma\) generally yield better accuracy and formatting, with \(\gamma = 0.1\) performing best. When filtering is removed, however, both metrics collapse despite using the same \(\gamma\), confirming that filtering high-entropy completions is essential for stability. The entropy curves further clarify this behavior. Without filtering, entropy continually rises above \(\ln 2\), destabilizing formatting first and accuracy later. With filtering, entropy stays below \(\ln 2\) and decreases smoothly. Moderate entropy (achieved with larger \(\gamma\)) supports exploration, producing more diverse and informative completions; excessively low entropy (e.g., with \(\gamma = 0.001\)) leads to overconfidence and early performance saturation. Finally, note that larger \(\gamma\) values tend to maintain slightly higher entropy. When a completion receives a negative advantage, entropy naturally increases (Eq.~\ref{eq:first_formula}), and a stronger regularization term (Eq.~\ref{eq:final-our}) amplifies this effect by slightly flattening the token distribution. However, this increase remains controlled and does not destabilize training; instead, it preserves a small amount of exploration that remains beneficial even in later stages.

\noindent \textbf{\(\langle H \rangle_{\text{i}}\) Threshold in GTPO.}
Figure~\ref{fig:ablation_ini_llama} analyzes the effect of different entropy thresholds \(\langle H \rangle_{\text{i}}\) on GTPO training for LLaMA on GSM8K. We consider three filtering settings: \(0.7\), \(1.05\), and \(2.0\), which roughly correspond to output distributions over, respectively, 2, 3, and 4 equiprobable tokens (with no residual probability mass). With the stricter thresholds (0.7 and 1.05), the entropy plot shows that the average completion entropy remains low and stable, while the loose threshold (2.0) lets many high-entropy completions pass, causing entropy to drift toward the instability region around \(\ln 2\) and destabilizing the policy. The accuracy curves mirror this behavior: tighter thresholds yield smoother improvements and higher final accuracy, whereas the loose threshold leads to stagnation. The formatting plot further confirms that insufficient filtering (as in the 2.0 setting) quickly degrades structural quality, whereas stricter thresholds preserve formatting and thus support both stability and output quality.

\noindent \textbf{Effects of Negative-Only Entropy Filtering.}
Figure \ref{fig:ablation_only_neg} compares GRPO against GTPO variants using Qwen on GSM8K dataset, analyzing different entropy filtering strategies. In \emph{OnlyNeg} setting, the entropy threshold, tested at \(\langle H \rangle_{\text{i}}< 2.00\), \(< 1.05\) and $< 0.7$, is applied only to completions with negative advantages, meaning all positive completions are retained. Instead the standard approach for Qwen (pink curve) does not apply the completion filtering. The results show that restricting filtering only to negative completions worsens performance; it drives entropy too low, reducing exploration, while failing to filter high entropy positive completions that may reinforce erroneous trajectories. 

\noindent \textbf{Effect of Regularization and Filtering on GRPO.}
Figure~\ref{fig:ablation_grpo_our} compares LLaMA on GSM8K under three settings: (i) standard GRPO without KL (\(\beta = 0\)), (ii) GRPO with entropy filtering (\(\langle H \rangle_i < 1.05\)) and entropy regularization (\(\gamma = 0.1\)), and (iii) GTPO with the same entropy constraints. The entropy plot (left) shows that standard GRPO without KL is highly unstable: entropy quickly spikes and then collapses, leading to a complete loss of structure. This is also reflected in the performance curves, where accuracy remains low and formatting breaks down early. Adding entropy regularization and filtering (roughly corresponding to three equiprobable tokens) significantly improves the situation. The entropy stays within a stable range, accuracy rises to reasonable levels, and formatting recovers. This shows that entropy control alone does not fully fix GRPO, but it substantially improves its stability and prevents the immediate collapse observed in the baseline. However, GTPO performs consistently better even under the same constraints. It maintains a lower and more stable entropy, achieves higher final accuracy, and preserves near-perfect formatting throughout training. In other words, while entropy control stabilizes GRPO, GTPO provides an additional layer of robustness and leads to the best overall performance.

\section{Conclusion}
\label{sec:conclusion}


In this work, we introduced \textit{Group-relative Trajectory-based Policy Optimization} (GTPO), which addresses two issues of GRPO, gradient conflicts and policy collapse, through conflict-aware updates and entropy control. We evaluated GTPO using LLaMA~8B and Qwen~2.5~(3B) on mathematical reasoning tasks, covering in-distribution datasets (GSM8K, MATH) and out-of-distribution benchmarks (AIME 2024, AIME 2025, AMC~2023). Through extensive ablation studies, we analyzed hyperparameter sensitivity to show the impact of each GTPO Objective component and how some of these components (entropy regularization and completion filtering) can also improve the performances of GRPO. On in-ditribution datasets, our results demonstrate that GTPO significantly outperforms baselines, improving Pass@32 and Maj@32 by up to 10 percentage points over GRPO and SFT. On out-of-distribution benchmarks, the gains are even larger, up to 75 points on Pass@64 and 37.5 points on Maj@64 over GRPO, and 32.5 and 42.5 points over SFT. For future work, we plan to further identify theoretical entropy bounds to ensure effective exploration without compromising stability.

\bibliography{bibliography}

@article{liu2025understanding,
  title={Understanding r1-zero-like training: A critical perspective},
  author={Liu, Zichen and Chen, Changyu and Li, Wenjun and Qi, Penghui and Pang, Tianyu and Du, Chao and Lee, Wee Sun and Lin, Min},
  journal={arXiv preprint arXiv:2503.20783},
  year={2025}
}

@article{zheng2025group,
  title={Group sequence policy optimization},
  author={Zheng, Chujie and Liu, Shixuan and Li, Mingze and Chen, Xiong-Hui and Yu, Bowen and Gao, Chang and Dang, Kai and Liu, Yuqiong and Men, Rui and Yang, An and others},
  journal={arXiv preprint arXiv:2507.18071},
  year={2025}
}

@article{hu2025open,
  title={Open-reasoner-zero: An open source approach to scaling up reinforcement learning on the base model},
  author={Hu, Jingcheng and Zhang, Yinmin and Han, Qi and Jiang, Daxin and Zhang, Xiangyu and Shum, Heung-Yeung},
  journal={arXiv preprint arXiv:2503.24290},
  year={2025}
}

@inproceedings{rlhf,
  author       = {Long Ouyang and
                  Jeffrey Wu and
                  Xu Jiang and
                  Diogo Almeida and
                  Carroll L. Wainwright and
                  Pamela Mishkin and
                  Chong Zhang and
                  Sandhini Agarwal and
                  Katarina Slama and
                  Alex Ray and
                  John Schulman and
                  Jacob Hilton and
                  Fraser Kelton and
                  Luke Miller and
                  Maddie Simens and
                  Amanda Askell and
                  Peter Welinder and
                  Paul F. Christiano and
                  Jan Leike and
                  Ryan Lowe},
  editor       = {Sanmi Koyejo and
                  S. Mohamed and
                  A. Agarwal and
                  Danielle Belgrave and
                  K. Cho and
                  A. Oh},
  title        = {Training language models to follow instructions with human feedback},
  booktitle    = {Advances in Neural Information Processing Systems 35: Annual Conference
                  on Neural Information Processing Systems 2022, NeurIPS 2022, New Orleans,
                  LA, USA, November 28 - December 9, 2022},
  year         = {2022},
  url          = {http://papers.nips.cc/paper\_files/paper/2022/hash/b1efde53be364a73914f58805a001731-Abstract-Conference.html},
  bibsource    = {dblp computer science bibliography, https://dblp.org}
}

@article{rlhf-notLLM,
  author       = {Volodymyr Mnih and
                  Koray Kavukcuoglu and
                  David Silver and
                  Andrei A. Rusu and
                  Joel Veness and
                  Marc G. Bellemare and
                  Alex Graves and
                  Martin A. Riedmiller and
                  Andreas Fidjeland and
                  Georg Ostrovski and
                  Stig Petersen and
                  Charles Beattie and
                  Amir Sadik and
                  Ioannis Antonoglou and
                  Helen King and
                  Dharshan Kumaran and
                  Daan Wierstra and
                  Shane Legg and
                  Demis Hassabis},
  title        = {Human-level control through deep reinforcement learning},
  journal      = {Nat.},
  volume       = {518},
  number       = {7540},
  pages        = {529--533},
  year         = {2015},
  url          = {https://doi.org/10.1038/nature14236},
  doi          = {10.1038/NATURE14236},
  bibsource    = {dblp computer science bibliography, https://dblp.org}
}

@inproceedings{google-rl,
  author       = {Volodymyr Mnih and
                  Adri{\`{a}} Puigdom{\`{e}}nech Badia and
                  Mehdi Mirza and
                  Alex Graves and
                  Timothy P. Lillicrap and
                  Tim Harley and
                  David Silver and
                  Koray Kavukcuoglu},
  editor       = {Maria{-}Florina Balcan and
                  Kilian Q. Weinberger},
  title        = {Asynchronous Methods for Deep Reinforcement Learning},
  booktitle    = {Proceedings of the 33nd International Conference on Machine Learning,
                  {ICML} 2016, New York City, NY, USA, June 19-24, 2016},
  series       = {{JMLR} Workshop and Conference Proceedings},
  volume       = {48},
  pages        = {1928--1937},
  publisher    = {JMLR.org},
  year         = {2016},
  url          = {http://proceedings.mlr.press/v48/mniha16.html},
  bibsource    = {dblp computer science bibliography, https://dblp.org}
}

@article{dota-2,
  author       = {Christopher Berner and
                  Greg Brockman and
                  Brooke Chan and
                  Vicki Cheung and
                  Przemyslaw Debiak and
                  Christy Dennison and
                  David Farhi and
                  Quirin Fischer and
                  Shariq Hashme and
                  Christopher Hesse and
                  Rafal J{\'{o}}zefowicz and
                  Scott Gray and
                  Catherine Olsson and
                  Jakub Pachocki and
                  Michael Petrov and
                  Henrique Pond{\'{e}} de Oliveira Pinto and
                  Jonathan Raiman and
                  Tim Salimans and
                  Jeremy Schlatter and
                  Jonas Schneider and
                  Szymon Sidor and
                  Ilya Sutskever and
                  Jie Tang and
                  Filip Wolski and
                  Susan Zhang},
  title        = {Dota 2 with Large Scale Deep Reinforcement Learning},
  journal      = {CoRR},
  volume       = {abs/1912.06680},
  year         = {2019},
  url          = {http://arxiv.org/abs/1912.06680},
  eprinttype    = {arXiv},
  eprint       = {1912.06680},
  bibsource    = {dblp computer science bibliography, https://dblp.org}
}

@article{GRPO-r1,
  title={Deepseek-r1: Incentivizing reasoning capability in llms via reinforcement learning},
  author={Guo, Daya and Yang, Dejian and Zhang, Haowei and Song, Junxiao and Zhang, Ruoyu and Xu, Runxin and Zhu, Qihao and Ma, Shirong and Wang, Peiyi and Bi, Xiao and others},
  journal={arXiv preprint arXiv:2501.12948},
  year={2025}
}

@article{GRPO,
  title={Deepseekmath: Pushing the limits of mathematical reasoning in open language models},
  author={Shao, Zhihong and Wang, Peiyi and Zhu, Qihao and Xu, Runxin and Song, Junxiao and Bi, Xiao and Zhang, Haowei and Zhang, Mingchuan and Li, YK and Wu, Yang and others},
  journal={arXiv preprint arXiv:2402.03300},
  year={2024}
}

@article{antropic-rlhf,
  author       = {Yuntao Bai and
                  Andy Jones and
                  Kamal Ndousse and
                  Amanda Askell and
                  Anna Chen and
                  Nova DasSarma and
                  Dawn Drain and
                  Stanislav Fort and
                  Deep Ganguli and
                  Tom Henighan and
                  Nicholas Joseph and
                  Saurav Kadavath and
                  Jackson Kernion and
                  Tom Conerly and
                  Sheer El Showk and
                  Nelson Elhage and
                  Zac Hatfield{-}Dodds and
                  Danny Hernandez and
                  Tristan Hume and
                  Scott Johnston and
                  Shauna Kravec and
                  Liane Lovitt and
                  Neel Nanda and
                  Catherine Olsson and
                  Dario Amodei and
                  Tom B. Brown and
                  Jack Clark and
                  Sam McCandlish and
                  Chris Olah and
                  Benjamin Mann and
                  Jared Kaplan},
  title        = {Training a Helpful and Harmless Assistant with Reinforcement Learning
                  from Human Feedback},
  journal      = {CoRR},
  volume       = {abs/2204.05862},
  year         = {2022},
  url          = {https://doi.org/10.48550/arXiv.2204.05862},
  doi          = {10.48550/ARXIV.2204.05862},
  eprinttype    = {arXiv},
  eprint       = {2204.05862},
  bibsource    = {dblp computer science bibliography, https://dblp.org}
}

@techreport{anthropic2024claude3,
  author       = {{Anthropic}},
  title        = {The Claude 3 Model Family: Opus, Sonnet, Haiku — Model Card},
  institution  = {Anthropic},
  type         = {Model Card},
  year         = {2024},
  month        = mar,
  day          = 4,
  url          = {https://assets.anthropic.com/m/61e7d27f8c8f5919/original/Claude-3-Model-Card.pdf},
  note         = {Accessed: 2025-07-22}
}

@article{gemini,
  author       = {Rohan Anil and
                  Sebastian Borgeaud and
                  Yonghui Wu and
                  Jean{-}Baptiste Alayrac and
                  Jiahui Yu and
                  Radu Soricut and
                  Johan Schalkwyk and
                  Andrew M. Dai and
                  Anja Hauth and
                  Katie Millican and
                  David Silver and
                  Slav Petrov and
                  Melvin Johnson and
                  Ioannis Antonoglou and
                  Julian Schrittwieser and
                  Amelia Glaese and
                  Jilin Chen and
                  Emily Pitler and
                  Timothy P. Lillicrap and
                  Angeliki Lazaridou and
                  Orhan Firat and
                  James Molloy and
                  Michael Isard and
                  Paul Ronald Barham and
                  Tom Hennigan and
                  Benjamin Lee and
                  Fabio Viola and
                  Malcolm Reynolds and
                  Yuanzhong Xu and
                  Ryan Doherty and
                  Eli Collins and
                  Clemens Meyer and
                  Eliza Rutherford and
                  Erica Moreira and
                  Kareem Ayoub and
                  Megha Goel and
                  George Tucker and
                  Enrique Piqueras and
                  Maxim Krikun and
                  Iain Barr and
                  Nikolay Savinov and
                  Ivo Danihelka and
                  Becca Roelofs and
                  Ana{\"{\i}}s White and
                  Anders Andreassen and
                  Tamara von Glehn and
                  Lakshman Yagati and
                  Mehran Kazemi and
                  Lucas Gonzalez and
                  Misha Khalman and
                  Jakub Sygnowski and
                  et al.},
  title        = {Gemini: {A} Family of Highly Capable Multimodal Models},
  journal      = {CoRR},
  volume       = {abs/2312.11805},
  year         = {2023},
  url          = {https://doi.org/10.48550/arXiv.2312.11805},
  doi          = {10.48550/ARXIV.2312.11805},
  eprinttype    = {arXiv},
  eprint       = {2312.11805},
  bibsource    = {dblp computer science bibliography, https://dblp.org}
}

@article{GPT4,
  author       = {OpenAI},
  title        = {{GPT-4} Technical Report},
  journal      = {CoRR},
  volume       = {abs/2303.08774},
  year         = {2023},
  url          = {https://doi.org/10.48550/arXiv.2303.08774},
  doi          = {10.48550/ARXIV.2303.08774},
  eprinttype    = {arXiv},
  eprint       = {2303.08774},
  bibsource    = {dblp computer science bibliography, https://dblp.org}
}

@inproceedings{TRPO,
  author       = {John Schulman and
                  Sergey Levine and
                  Pieter Abbeel and
                  Michael I. Jordan and
                  Philipp Moritz},
  editor       = {Francis R. Bach and
                  David M. Blei},
  title        = {Trust Region Policy Optimization},
  booktitle    = {Proceedings of the 32nd International Conference on Machine Learning,
                  {ICML} 2015, Lille, France, 6-11 July 2015},
  series       = {{JMLR} Workshop and Conference Proceedings},
  volume       = {37},
  pages        = {1889--1897},
  publisher    = {JMLR.org},
  year         = {2015},
  url          = {http://proceedings.mlr.press/v37/schulman15.html},
  bibsource    = {dblp computer science bibliography, https://dblp.org}
}

@inproceedings{ppo37implementations,
  author = {Huang, Shengyi and Dossa, Rousslan Fernand Julien and Raffin, Antonin and Kanervisto, Anssi and Wang, Weixun},
  title = {The 37 Implementation Details of Proximal Policy Optimization},
  booktitle = {ICLR Blog Track},
  year = {2022},
  note = {https://iclr-blog-track.github.io/2022/03/25/ppo-implementation-details/},
  url  = {https://iclr-blog-track.github.io/2022/03/25/ppo-implementation-details/}
}

@inproceedings{trgppo,
  author       = {Yuhui Wang and
                  Hao He and
                  Xiaoyang Tan and
                  Yaozhong Gan},
  editor       = {Hanna M. Wallach and
                  Hugo Larochelle and
                  Alina Beygelzimer and
                  Florence d'Alch{\'{e}}{-}Buc and
                  Emily B. Fox and
                  Roman Garnett},
  title        = {Trust Region-Guided Proximal Policy Optimization},
  booktitle    = {Advances in Neural Information Processing Systems 32: Annual Conference
                  on Neural Information Processing Systems 2019, NeurIPS 2019, December
                  8-14, 2019, Vancouver, BC, Canada},
  pages        = {624--634},
  year         = {2019},
  url          = {https://proceedings.neurips.cc/paper/2019/hash/a666587afda6e89aec274a3657558a27-Abstract.html},
  bibsource    = {dblp computer science bibliography, https://dblp.org}
}

@inproceedings{PPOissues,
  author       = {Skander Moalla and
                  Andrea Miele and
                  Daniil Pyatko and
                  Razvan Pascanu and
                  Caglar Gulcehre},
  editor       = {Amir Globersons and
                  Lester Mackey and
                  Danielle Belgrave and
                  Angela Fan and
                  Ulrich Paquet and
                  Jakub M. Tomczak and
                  Cheng Zhang},
  title        = {No Representation, No Trust: Connecting Representation, Collapse,
                  and Trust Issues in {PPO}},
  booktitle    = {Advances in Neural Information Processing Systems 38: Annual Conference
                  on Neural Information Processing Systems 2024, NeurIPS 2024, Vancouver,
                  BC, Canada, December 10 - 15, 2024},
  year         = {2024},
  url          = {http://papers.nips.cc/paper\_files/paper/2024/hash/81166fbd9cc5adf14031cdb69d3fd6a8-Abstract-Conference.html},
  bibsource    = {dblp computer science bibliography, https://dblp.org}
}

@inproceedings{PPOissues2,
  author       = {Saurabh Garg and
                  Joshua Zhanson and
                  Emilio Parisotto and
                  Adarsh Prasad and
                  J. Zico Kolter and
                  Zachary C. Lipton and
                  Sivaraman Balakrishnan and
                  Ruslan Salakhutdinov and
                  Pradeep Ravikumar},
  editor       = {Marina Meila and
                  Tong Zhang},
  title        = {On Proximal Policy Optimization's Heavy-tailed Gradients},
  booktitle    = {Proceedings of the 38th International Conference on Machine Learning,
                  {ICML} 2021, 18-24 July 2021, Virtual Event},
  series       = {Proceedings of Machine Learning Research},
  volume       = {139},
  pages        = {3610--3619},
  publisher    = {{PMLR}},
  year         = {2021},
  url          = {http://proceedings.mlr.press/v139/garg21b.html},
  bibsource    = {dblp computer science bibliography, https://dblp.org}
}

@article{alphaPPO,
  author       = {Haotian Xu and
                  Zheng Yan and
                  Junyu Xuan and
                  Guangquan Zhang and
                  Jie Lu},
  title        = {Improving proximal policy optimization with alpha divergence},
  journal      = {Neurocomputing},
  volume       = {534},
  pages        = {94--105},
  year         = {2023},
  url          = {https://doi.org/10.1016/j.neucom.2023.02.008},
  doi          = {10.1016/J.NEUCOM.2023.02.008},
  bibsource    = {dblp computer science bibliography, https://dblp.org}
}

@article{PPO-ALR,
  author       = {Lu Jia and
                  Binglin Su and
                  Du Xu and
                  Yewei Wang and
                  Jing Fang and
                  Jun Wang},
  title        = {Policy Optimization Algorithm with Activation Likelihood-Ratio for
                  Multi-agent Reinforcement Learning},
  journal      = {Neural Process. Lett.},
  volume       = {56},
  number       = {6},
  pages        = {247},
  year         = {2024},
  url          = {https://doi.org/10.1007/s11063-024-11705-x},
  doi          = {10.1007/S11063-024-11705-X},
  bibsource    = {dblp computer science bibliography, https://dblp.org}
}

@inproceedings{GSM8K,
  author       = {Dan Hendrycks and
                  Collin Burns and
                  Saurav Kadavath and
                  Akul Arora and
                  Steven Basart and
                  Eric Tang and
                  Dawn Song and
                  Jacob Steinhardt},
  editor       = {Joaquin Vanschoren and
                  Sai{-}Kit Yeung},
  title        = {Measuring Mathematical Problem Solving With the {MATH} Dataset},
  booktitle    = {Proceedings of the Neural Information Processing Systems Track on
                  Datasets and Benchmarks 1, NeurIPS Datasets and Benchmarks 2021, December
                  2021, virtual},
  year         = {2021},
}

@inproceedings{MATH,
  author       = {Dan Hendrycks and
                  Collin Burns and
                  Saurav Kadavath and
                  Akul Arora and
                  Steven Basart and
                  Eric Tang and
                  Dawn Song and
                  Jacob Steinhardt},
  editor       = {Joaquin Vanschoren and
                  Sai{-}Kit Yeung},
  title        = {Measuring Mathematical Problem Solving With the {MATH} Dataset},
  booktitle    = {Proceedings of the Neural Information Processing Systems Track on
                  Datasets and Benchmarks 1, NeurIPS Datasets and Benchmarks 2021, December
                  2021, virtual},
  year         = {2021},
}

@article{simoni2025improving,
  title={Improving LLM Reasoning for Vulnerability Detection via Group Relative Policy Optimization},
  author={Simoni, Marco and Fontana, Aleksandar and Rossolini, Giulio and Saracino, Andrea},
  journal={arXiv preprint arXiv:2507.03051},
  year={2025}
}

@misc{cui2025entropymechanismreinforcementlearning,
      title={The Entropy Mechanism of Reinforcement Learning for Reasoning Language Models}, 
      author={Ganqu Cui and Yuchen Zhang and Jiacheng Chen and Lifan Yuan and Zhi Wang and Yuxin Zuo and Haozhan Li and Yuchen Fan and Huayu Chen and Weize Chen and Zhiyuan Liu and Hao Peng and Lei Bai and Wanli Ouyang and Yu Cheng and Bowen Zhou and Ning Ding},
      year={2025},
      eprint={2505.22617},
      archivePrefix={arXiv},
      primaryClass={cs.LG},
      url={https://arxiv.org/abs/2505.22617}, 
}

@software{unsloth,
  author = {Daniel Han, Michael Han and Unsloth team},
  title = {Unsloth},
  url = {http://github.com/unslothai/unsloth},
  year = {2023}
}

@inproceedings{dohare2023overcoming,
  title={Overcoming policy collapse in deep reinforcement learning},
  author={Dohare, Shibhansh and Lan, Qingfeng and Mahmood, A Rupam},
  booktitle={Sixteenth European Workshop on Reinforcement Learning},
  year={2023}
}

@article{ppo,
  title={Proximal policy optimization algorithms},
  author={Schulman, John and Wolski, Filip and Dhariwal, Prafulla and Radford, Alec and Klimov, Oleg},
  journal={arXiv preprint arXiv:1707.06347},
  year={2017}
}

@article{llama,
  title={The carbon footprint of machine learning training will plateau, then shrink},
  author={Patterson, David and Gonzalez, Joseph and H{\"o}lzle, Urs and Le, Quoc and Liang, Chen and Munguia, Lluis-Miquel and Rothchild, Daniel and So, David R and Texier, Maud and Dean, Jeff},
  journal={Computer},
  volume={55},
  number={7},
  pages={18--28},
  year={2022},
  publisher={IEEE}
}

@article{qwen,
  title={Qwen3 technical report},
  author={Yang, An and Li, Anfeng and Yang, Baosong and Zhang, Beichen and Hui, Binyuan and Zheng, Bo and Yu, Bowen and Gao, Chang and Huang, Chengen and Lv, Chenxu and others},
  journal={arXiv preprint arXiv:2505.09388},
  year={2025}
}

@misc{RFT,
      title={Scaling Relationship on Learning Mathematical Reasoning with Large Language Models}, 
      author={Zheng Yuan and Hongyi Yuan and Chengpeng Li and Guanting Dong and Keming Lu and Chuanqi Tan and Chang Zhou and Jingren Zhou},
      year={2023},
      eprint={2308.01825},
      archivePrefix={arXiv},
      primaryClass={cs.CL},
      url={https://arxiv.org/abs/2308.01825}, 
}

@inproceedings{DPO,
  author       = {Rafael Rafailov and
                  Archit Sharma and
                  Eric Mitchell and
                  Christopher D. Manning and
                  Stefano Ermon and
                  Chelsea Finn},
  editor       = {Alice Oh and
                  Tristan Naumann and
                  Amir Globerson and
                  Kate Saenko and
                  Moritz Hardt and
                  Sergey Levine},
  title        = {Direct Preference Optimization: Your Language Model is Secretly a
                  Reward Model},
  booktitle    = {Advances in Neural Information Processing Systems 36: Annual Conference
                  on Neural Information Processing Systems 2023, NeurIPS 2023, New Orleans,
                  LA, USA, December 10 - 16, 2023},
  year         = {2023},
  url          = {http://papers.nips.cc/paper\_files/paper/2023/hash/a85b405ed65c6477a4fe8302b5e06ce7-Abstract-Conference.html},
  bibsource    = {dblp computer science bibliography, https://dblp.org}
}

@article{grpo-alignment-human,
  author       = {Xuying Li and
                  Zhuo Li and
                  Yuji Kosuga and
                  Victor Bian},
  title        = {Optimizing Safe and Aligned Language Generation: {A} Multi-Objective
                  {GRPO} Approach},
  journal      = {CoRR},
  volume       = {abs/2503.21819},
  year         = {2025},
  url          = {https://doi.org/10.48550/arXiv.2503.21819},
  doi          = {10.48550/ARXIV.2503.21819},
  eprinttype    = {arXiv},
  eprint       = {2503.21819},
  bibsource    = {dblp computer science bibliography, https://dblp.org}
}

@article{overfit-grpo,
  author       = {Andre He and
                  Daniel Fried and
                  Sean Welleck},
  title        = {Rewarding the Unlikely: Lifting {GRPO} Beyond Distribution Sharpening},
  journal      = {CoRR},
  volume       = {abs/2506.02355},
  year         = {2025},
  url          = {https://doi.org/10.48550/arXiv.2506.02355},
  doi          = {10.48550/ARXIV.2506.02355},
  eprinttype    = {arXiv},
  eprint       = {2506.02355},
  bibsource    = {dblp computer science bibliography, https://dblp.org}
}

@article{low-token-grpo-issue,
  author       = {Zhihe Yang and
                  Xufang Luo and
                  Zilong Wang and
                  Dongqi Han and
                  Zhiyuan He and
                  Dongsheng Li and
                  Yunjian Xu},
  title        = {Do Not Let Low-Probability Tokens Over-Dominate in {RL} for LLMs},
  journal      = {CoRR},
  volume       = {abs/2505.12929},
  year         = {2025},
  url          = {https://doi.org/10.48550/arXiv.2505.12929},
  doi          = {10.48550/ARXIV.2505.12929},
  eprinttype    = {arXiv},
  eprint       = {2505.12929},
  bibsource    = {dblp computer science bibliography, https://dblp.org}
}

@inproceedings{entropy-on-ppo,
  author       = {Marcin Andrychowicz and
                  Anton Raichuk and
                  Piotr Stanczyk and
                  Manu Orsini and
                  Sertan Girgin and
                  Rapha{\"{e}}l Marinier and
                  L{\'{e}}onard Hussenot and
                  Matthieu Geist and
                  Olivier Pietquin and
                  Marcin Michalski and
                  Sylvain Gelly and
                  Olivier Bachem},
  title        = {What Matters for On-Policy Deep Actor-Critic Methods? {A} Large-Scale
                  Study},
  booktitle    = {9th International Conference on Learning Representations, {ICLR} 2021,
                  Virtual Event, Austria, May 3-7, 2021},
  publisher    = {OpenReview.net},
  year         = {2021},
  url          = {https://openreview.net/forum?id=nIAxjsniDzg},
  bibsource    = {dblp computer science bibliography, https://dblp.org}
}

@article{pass@k,
  author       = {Mark Chen and
                  Jerry Tworek and
                  Heewoo Jun and
                  Qiming Yuan and
                  Henrique Pond{\'{e}} de Oliveira Pinto and
                  Jared Kaplan and
                  Harri Edwards and
                  Yuri Burda and
                  Nicholas Joseph and
                  Greg Brockman and
                  Alex Ray and
                  Raul Puri and
                  Gretchen Krueger and
                  Michael Petrov and
                  Heidy Khlaaf and
                  Girish Sastry and
                  Pamela Mishkin and
                  Brooke Chan and
                  Scott Gray and
                  Nick Ryder and
                  Mikhail Pavlov and
                  Alethea Power and
                  Lukasz Kaiser and
                  Mohammad Bavarian and
                  Clemens Winter and
                  Philippe Tillet and
                  Felipe Petroski Such and
                  Dave Cummings and
                  Matthias Plappert and
                  Fotios Chantzis and
                  Elizabeth Barnes and
                  Ariel Herbert{-}Voss and
                  William Hebgen Guss and
                  Alex Nichol and
                  Alex Paino and
                  Nikolas Tezak and
                  Jie Tang and
                  Igor Babuschkin and
                  Suchir Balaji and
                  Shantanu Jain and
                  William Saunders and
                  Christopher Hesse and
                  Andrew N. Carr and
                  Jan Leike and
                  Joshua Achiam and
                  Vedant Misra and
                  Evan Morikawa and
                  Alec Radford and
                  Matthew Knight and
                  Miles Brundage and
                  Mira Murati and
                  Katie Mayer and
                  Peter Welinder and
                  Bob McGrew and
                  Dario Amodei and
                  Sam McCandlish and
                  Ilya Sutskever and
                  Wojciech Zaremba},
  title        = {Evaluating Large Language Models Trained on Code},
  journal      = {CoRR},
  volume       = {abs/2107.03374},
  year         = {2021},
  url          = {https://arxiv.org/abs/2107.03374},
  eprinttype    = {arXiv},
  eprint       = {2107.03374},
  bibsource    = {dblp computer science bibliography, https://dblp.org}
}

@inproceedings{maj@k,
  author       = {Xuezhi Wang and
                  Jason Wei and
                  Dale Schuurmans and
                  Quoc V. Le and
                  Ed H. Chi and
                  Sharan Narang and
                  Aakanksha Chowdhery and
                  Denny Zhou},
  title        = {Self-Consistency Improves Chain of Thought Reasoning in Language Models},
  booktitle    = {The Eleventh International Conference on Learning Representations,
                  {ICLR} 2023, Kigali, Rwanda, May 1-5, 2023},
  publisher    = {OpenReview.net},
  year         = {2023},
  url          = {https://openreview.net/forum?id=1PL1NIMMrw},
  bibsource    = {dblp computer science bibliography, https://dblp.org}
}

@article{aime2024,
  title={Numinamath: The largest public dataset in ai4maths with 860k pairs of competition math problems and solutions},
  author={Li, Jia and Beeching, Edward and Tunstall, Lewis and Lipkin, Ben and Soletskyi, Roman and Huang, Shengyi and Rasul, Kashif and Yu, Longhui and Jiang, Albert Q and Shen, Ziju and others},
  journal={Hugging Face repository},
  volume={13},
  number={9},
  pages={9},
  year={2024}
}

@article{aime2025,
  title={Numinamath: The largest public dataset in ai4maths with 860k pairs of competition math problems and solutions},
  author={Li, Jia and Beeching, Edward and Tunstall, Lewis and Lipkin, Ben and Soletskyi, Roman and Huang, Shengyi and Rasul, Kashif and Yu, Longhui and Jiang, Albert Q and Shen, Ziju and others},
  journal={Hugging Face repository},
  volume={13},
  number={9},
  pages={9},
  year={2024}
}

@article{amc2023,
  title={Numinamath: The largest public dataset in ai4maths with 860k pairs of competition math problems and solutions},
  author={Li, Jia and Beeching, Edward and Tunstall, Lewis and Lipkin, Ben and Soletskyi, Roman and Huang, Shengyi and Rasul, Kashif and Yu, Longhui and Jiang, Albert Q and Shen, Ziju and others},
  journal={Hugging Face repository},
  volume={13},
  number={9},
  pages={9},
  year={2024}
}

@inproceedings{first-token-similar,
  author       = {Xinpeng Wang and
                  Bolei Ma and
                  Chengzhi Hu and
                  Leon Weber{-}Genzel and
                  Paul R{\"{o}}ttger and
                  Frauke Kreuter and
                  Dirk Hovy and
                  Barbara Plank},
  editor       = {Lun{-}Wei Ku and
                  Andre Martins and
                  Vivek Srikumar},
  title        = {"My Answer is C": First-Token Probabilities Do Not Match Text Answers
                  in Instruction-Tuned Language Models},
  booktitle    = {Findings of the Association for Computational Linguistics, {ACL} 2024,
                  Bangkok, Thailand and virtual meeting, August 11-16, 2024},
  pages        = {7407--7416},
  publisher    = {Association for Computational Linguistics},
  year         = {2024},
  url          = {https://doi.org/10.18653/v1/2024.findings-acl.441},
  doi          = {10.18653/V1/2024.FINDINGS-ACL.441},
  bibsource    = {dblp computer science bibliography, https://dblp.org}
}
\bibliographystyle{acl_natbib}
\clearpage

\twocolumn[
  \begin{@twocolumnfalse}
\begin{center}
  \Large
  \textbf{Appendix}
\end{center} 
\vspace{2em}

\end{@twocolumnfalse}
]





\section{Proofs and Repricability}
\subsection{Group-Relative Policy Optimization}
\par \textbf{The Role of KL Divergence on GRPO.}
\label{appendix:aggregated-grpo-loss} We analyze the impact of the KL divergence term on the aggregated GRPO loss by transitioning from a token-level formulation to a trajectory-level one, when $r_{i,t}=1$. 
Recall the normalized advantage function used in GRPO-like:

\begin{equation}
    \hat{\mathcal{A}}_{i,t} = \hat{\mathcal{A}}_i = \frac{R_i - \bar{R}}{\operatorname{std}(R)},
    \label{eq:normalized-advantage}
\end{equation}

where $R_i$ is the scalar reward associated with trajectory $i$, $\bar{R}$ is the average reward across all $G$ trajectories in the batch, and $\operatorname{std}(R)$ denotes the standard deviation of the rewards.

Substituting Equation~\ref{eq:normalized-advantage} into the GRPO objective from Equation~\ref{eq:grpo-full}, we obtain:

\footnotesize
\begin{align}
    \mathcal{J}_{\mathrm{GRPO}}(\theta) & = \mathbb{E}_{q,\{o_i\}} \Bigg[\frac{1}{G} \sum_{i=1}^{G}\frac{1}{|o_i|} \sum_{t=1}^{|o_i|} \frac{R_i - \bar{R}}{\operatorname{std}(R)} \notag \\
    & - \beta \cdot \mathrm{D}_{\mathrm{KL}} \left( \pi_{\theta} \| \pi_{\mathrm{ref}} \right)\Bigg]
    \label{eq:grpo-loss-trajectory}
\end{align}
\normalsize

We now turn our attention to the first term of the loss, which is defined as follows:

\begin{equation}
    \tilde{\mathcal{J}}_{\mathrm{GRPO}}(\theta) := \mathbb{E}_{q,\{o_i\}} \Bigg[\frac{1}{G} \sum_{i=1}^{G} \frac{1}{|o_i|} \sum_{t=1}^{|o_i|}\frac{R_i - \bar{R}}{\operatorname{std}(R)} \Bigg]
    \label{eq:reward-term-grpo}
\end{equation}

Since equation \ref{eq:normalized-advantage} is independent of $t$, it is possible to simplify $\tilde{\mathcal{J}}_{\mathrm{GRPO}}(\theta)$ as follows:

\small
\begin{equation}
\begin{aligned}
    \tilde{\mathcal{J}}_{\mathrm{GRPO}}(\theta)
    &= \mathbb{E}_{q,\{o_i\}} \Bigg[\frac{1}{G} \sum_{i=1}^{G} \left( \frac{R_i}{\operatorname{std}(R)} - \frac{\bar{R}}{\operatorname{std}(R)} \right) \Bigg] \\
    &= \mathbb{E}_{q,\{o_i\}} \Bigg[\frac{1}{\operatorname{std}(R)} \left( \frac{1}{G} \sum_{i=1}^{G} R_i - \bar{R} \right) \Bigg]\\
    &= \mathbb{E}_{q,\{o_i\}} [0] .
\end{aligned}
    \label{eq:zero-mean-constrain}
\end{equation}

\normalsize

Thus, the overall GRPO objective reduces to:

\begin{equation}
    \mathcal{J}_{\text{GRPO}}(\theta) = \mathbb{E}_{q,\{o_i\}} \Big[- \beta \cdot \mathrm{D}_{\mathrm{KL}} \left( \pi_{\theta} \| \pi_{\mathrm{ref}} \right)\Big]
    \label{eq:grpo-loss-aggregated}
\end{equation}

From this aggregated viewpoint, the loss is entirely governed by the KL divergence term. However, it is crucial to emphasize that a zero-valued aggregated reward term \textit{does not imply that the gradient of the original GRPO loss is zero}. \textit{The per-token advantages still guide parameter updates during optimization, ensuring meaningful learning even when the aggregated advantage cancels out.}

\par \textbf{GRPO Gradient.}
\label{appendix:grpo-gradient} Given the GRPO objective defined in Equation~\ref{eq:grpo-full}, we aim to calculate the gradient. 

\footnotesize
\begin{equation}
\tilde{\mathcal{J}}_{\text{GRPO}}(\theta) = \mathbb{E}_{q,\{o_i\}} \Bigg[\frac{1}{G} \sum_{i=1}^{G} \frac{ \hat{\mathcal{A}}_{i}}{|o_i|} \sum_{t=1}^{|o_i|}
\frac{\pi_\theta(o_{i,t} \mid s_{i,t})}{\pi_{\theta_{\text{old}}}(o_{i,t} \mid s_{i,t})}\Bigg]
\label{eq:grpo-token-for-grad}
\end{equation}

\normalsize
To simplify the calculation, we keep $\hat{\mathcal{A}}_i$ in its unexpanded form (Eq. \ref{eq:normalized-advantage}) and reduce the term $C_{i,t}$,(Eq. \ref{eq:combined_objective}) to the ratio $r_{i,t}(\theta) = \frac{\pi_\theta(o_{i,t} \mid s_{i,t})}{\pi_{\theta_{\text{old}}}(o_{i,t} \mid s_{i,t})}$. This substitution relies on the fact that the objective function's clipping mechanism eliminates gradient contributions from tokens where $|r_{i,t}(\theta)|>1+\epsilon$. Therefore, the summation is performed exclusively over the set of active (non clipped) tokens. Formalizing the excluded tokens adds notational burden with no impact on the resulting gradient analysis.


\footnotesize
\begin{align}
\nabla_\theta &\tilde{\mathcal{J}}_{\text{GRPO}}(\theta) = \notag\\ & = \mathbb{E}_{q,\{o_i\}} \Bigg[\frac{1}{G} \sum_{i=1}^{G} \frac{ \hat{\mathcal{A}}_i}{|o_i|}  \sum_{t=1}^{|o_i|} \nabla_\theta \left[
 \frac{\pi_\theta(o_{i,t} \mid s_{i,t})}{\pi_{\theta_{\text{old}}}(o_{i,t} \mid s_{i,t})} \right]\Bigg] \notag \\
 & = \mathbb{E}_{q,\{o_i\}} \Bigg[\frac{1}{G} \sum_{i=1}^{G} \frac{ \hat{\mathcal{A}}_i}{|o_i|} \sum_{t=1}^{|o_i|} \frac{\pi_\theta(o_{i,t} \mid s_{i,t})}{\pi_{\theta_{\text{old}}}(o_{i,t} \mid s_{i,t})}\Bigg] \notag
\end{align}

\normalsize

We subsequently apply the log-derivative trick, a widely used technique in reinforcement learning, which reformulates the gradient as $\nabla_\theta \pi_\theta(x) = \pi_\theta(x) \nabla_\theta \log(\pi_\theta(x))$.

\footnotesize
\begin{align}
    & \nabla_\theta \tilde{\mathcal{J}}_{\text{GRPO}}(\theta) = \notag
    \\ &  \mathbb{E}_{q,\{o_i\}} \Bigg[\frac{1}{G} \sum_{i=1}^{G} \frac{ \hat{\mathcal{A}}{i}}{|o_i|} \sum_{t=1}^{|o_i|}  r_{i,t}(\theta)\cdot \nabla_\theta \left[\log(\pi_\theta(o_{i,t}|s_{i,t}))\right]\Bigg]
\end{align}

\normalsize
We can rewrite the complete gradient as:

\small
\begin{align}
 & \nabla_\theta  \mathcal{J}_{\text{GRPO}}(\theta) = \notag
 \\ & \mathbb{E}_{q,\{o_i\}} \Bigg[\frac{1}{G} \sum_{i=1}^{G} \frac{\hat{\mathcal{A}}_{i}}{|o_i|} \sum_{t=1}^{|o_i|} r_{i,t}(\theta)\nabla_\theta \left[\log(\pi_\theta(o_{i,t}|s_{i,t}))\right] \notag 
 \\ & - \beta \cdot \nabla_\theta \left[ \mathrm{D}_{\mathrm{KL}} \left( \pi_{\theta} \| \pi_{\mathrm{ref}} \right) \right] \Bigg]
\label{eq:grpo-complete-gradient}
\end{align}
\normalsize

\par \textbf{How gradients affect distribution over tokens.} LLMs typically terminate with a softmax layer, which produces a probability distribution over the vocabulary.  To simplify notation in what follows, we denote \( (o_{i,t}|q,o_{i,<t})\) as $(t')$, yielding the following expression for the output distribution:

\begin{equation}
\pi_\theta(t') = \frac{e^{f^j_\theta}}{\sum_k^V e^{f^k_\theta}}  
\label{eq:softmax}
\end{equation}

Here, $f^j_\theta$ denotes the chosen logit corresponding to position $o_{i,t}$, while $\sum_{k=1}^V f^k_\theta$ represents the aggregate logits over the entire vocabulary of size \( V \), conditioned on the context \( (q,o_{i,<t}) \).

Our objective is to compute the gradient in order to analyze its influence on the probability distribution of both the selected token and the unselected alternatives. 

\small
\begin{align}
\nabla_\theta \log(\pi_\theta(t')) & =     \nabla_\theta \log \left( \frac{e^{f^j_\theta}}{\sum_k^V e^{f^k_\theta}}   \right) \notag \\
& = \nabla_\theta \log \left( e^{f^j_\theta} \right) - \nabla_\theta \log \left( \sum_k^V e^{f^k_\theta}  \right)
\end{align}

\normalsize

The first term is just $\nabla_\theta e^{f^j_\theta}$; for the second, apply the chain rule:

\begin{align}
\nabla_\theta \log\left(\sum_k^V e^{f^k_\theta}\right) & =     \frac{1}{\sum_b^V e^{f^b_\theta}} \nabla_\theta \left( \sum_k^V e^{f^k_\theta}  \right) \notag \\
& = \sum_k^V \frac{e^{f^k_\theta}}{\sum_b^V e^{f^b_\theta}} \nabla_\theta f^k_\theta \notag \\
& = \sum_k^V \pi^k_\theta(t') \nabla_\theta f^k_\theta
\end{align}

Where $\pi^k_\theta(t')$ represents the probability of the possible token $k$ given the context $t'$, we can combine these as follows:
\small
\begin{align}
\nabla_\theta \log(\pi_\theta(t')) & = \nabla_\theta f^j_\theta - \sum_k^V \pi_\theta(k) \nabla_\theta f^k_\theta \notag \\
& = \nabla_\theta f^k_\theta (1-\pi_\theta(t')) - \sum_{k \neq t}^V \pi_\theta(k') \nabla_\theta f^k_\theta
\end{align}
\normalsize

Substituting this into the GRPO gradient yields:

\footnotesize

\begin{align}
\nabla_\theta & \mathcal{J}_{\text{GRPO}}(\theta) = \notag \\ & \mathbb{E}_{q,\{o_i\}} \Bigg[\frac{1}{G} \sum_{i=1}^{G} \frac{\hat{\mathcal{A}}_i}{|o_i|} \sum_{t=1}^{|o_i|} r_{i,t}\Bigg( \nabla_\theta f^k_\theta (1-\pi_\theta(t')) \notag 
 \\ & \left. - \sum_{k \neq t}^V \pi_\theta(k') \nabla_\theta f^k_\theta \right) -\beta \cdot \nabla_\theta \left[ \mathrm{D}_{\mathrm{KL}} \left( \pi_{\theta} \| \pi_{\mathrm{ref}} \right) \right] \Bigg]
\end{align}

\normalsize

\subsection{Group-relative Trajectory-aware Policy Optimization: Objective}
\label{sec:full-der-gtpo}
This section formalizes the \textit{Conflict-Aware Gradient Correction} for GTPO, deriving the token-level loss and gradient. GTPO refines GRPO by weighting each token with a coefficient $\lambda_{i,t}$, determined by conflict regions and the advantage $A_i$. Furthermore, to mitigate instability, we incorporate an entropy-based filter $\delta_i$. The resulting objective function is defined as:

\footnotesize
\begin{align*}
& \mathcal{J}^* := \mathbb{E}_{q,\{o_i\}} \Bigg[\frac{1}{G} \sum_{i=1}^G \frac{1}{|o_i|} \sum_{t \in o_i} r_{i,t}\lambda_{i,t} A_{i} \Bigg] \\
&= \mathbb{E}_{q,\{o_i\}} \Bigg[\frac{1}{G} \sum_{i=1}^G \frac{1}{|o_i|} \left( \sum_{t \in c_i} r_{i,t}\lambda_{i,t} A_{i} + \sum_{t \in u_i} r_{i,t}\lambda_{i,t} A_i \right) \Bigg] \\
\end{align*}

\normalsize

Where $o_i$ denotes the set of tokens in each completion, $c_i$ and $u_i$ represent the subsets of conflicting and non-conflicting tokens, respectively. The notation $|\cdot|$ indicates the cardinality (number of elements) of each set and $\lambda_{i,t}$ is a parameter designed to account for conflict and non-conflict formulation. To motivate the definition of $\lambda_{i,t}$, we assume the experimental condition where $r_{i,t}=1$. Furthermore, to ensure that the model retains the characteristics of GRPO for non-conflicting tokens, we define $\lambda_{i,t}$ as follows:

\footnotesize

$\lambda_{i,t} = \begin{cases}
\lambda_i, & \text{if t} \in c_i 
\\[4pt]
1, & \text{if t} \in u_i
\end{cases}
$

\begin{align*}
&\mathcal{J}^* 
= \mathbb{E}_{q,\{o_i\}} \Bigg[\frac{1}{G} \sum_{i=1}^G \frac{1}{|o_i|} \left( |c_i| \lambda_i A_{i} + \left(|o_i| - |c_i|\right) A_i \right) \Bigg] \\
&= \mathbb{E}_{q,\{o_i\}} \Bigg[\frac{1}{G} \left( \sum_{i=1}^G \frac{|c_i|}{|o_i|} \lambda_i A_{i} + \sum_{i=1}^G A_i - \sum_{i=1}^G \frac{|c_i|}{|o_i|} A_i \right) \Bigg]\\
&= \mathbb{E}_{q,\{o_i\}} \Bigg[\frac{1}{G} \left(\sum_{i=1}^G \frac{|c_i|}{|o_i|} A_i (\lambda_i -1) + \sum_{i=1}^G A_i \right) \Bigg]
\end{align*}

\normalsize

First of all, given the \textit{GRPO’s zero-mean constraint}, $\sum_i^G A_i = 0$, we split the sum into two parts: one considering only the positive completions ($G^+$), and the other considering only the negative ones ($G^-$).

\small

\begin{align*}
\mathcal{J}^* = \mathbb{E}_{q,\{o_i\}} \Bigg[&\frac{1}{G} \Bigg(\sum_{i=1}^{G^+} \frac{|c_i|}{|o_i|} A_i (\lambda_i -1) \\ & +\sum_{i=1}^{G^-} \frac{|c_i|}{|o_i|} A_i (\lambda_i -1) \Bigg)\Bigg]
\end{align*}

\normalsize

To prevent inadvertently penalizing tokens that contribute positively in other contexts, we refrain from assigning negative rewards to tokens involved in conflicting cases. These tokens frequently appear in completions that receive positive ratings and are not intrinsically responsible for the negative assessments.

$\lambda_i = \begin{cases}
    \lambda, &\text{if i} \in G^+\\
    0 &\text{if i} \in G^-
\end{cases}$

\footnotesize
\begin{align*}
\mathcal{J}^* 
&=\mathbb{E}_{q,\{o_i\}} \Bigg[\frac{1}{G} \left(\sum_{i=1}^{G^+} \frac{|c_i|}{|o_i|} |A_i| (\lambda -1) +\sum_{i=1}^{G^-} \frac{|c_i|}{|o_i|} |A_i| \right) \Bigg]\\
\end{align*}

\normalsize

It is evident that by setting $\lambda = 2$, we ensure equal contribution from each completion, regardless of whether it is positive or negative.

The resulting equation for the aggregated loss is:

\begin{align}
\mathcal{J}^* 
&=\mathbb{E}_{q,\{o_i\}} \Bigg[\frac{1}{G} \sum_{i=1}^{G} \frac{|c_i|}{|o_i|} |A_i| \Bigg]
\end{align}

Instead the resulting equation for the token level loss, considering $r_{i,t} \neq 1$, is:


\begin{align}
\mathcal{J}^* =& \mathbb{E}_{q,\{o_i\}} \Bigg[\frac{1}{G} \sum_{i=1}^{G} \frac{1}{|o_i|}
\sum_{t = 0}^{o_i}r_{i,t}\lambda_{i,t} A_i \Bigg]
\label{eq:our-token-level}
\end{align}

\noindent where:
\begin{equation}
\lambda_{i,t} = \begin{cases}
    1, & \text{if } i \in \{G^-, G^+\} \text{ and } t \in u_i \\
    0, & \text{if } i \in G^- \text{ and } t \in c_i \\
    2, & \text{if } i \in G^+ \text{ and } t \in c_i
\end{cases}
\end{equation}

\par \textbf{GTPO Gradient:}
To calculate the gradient of our proposed loss, we start from equation~\ref{eq:our-token-level}, then we apply the same tricks used to compute the GRPO loss (eq.~\ref{eq:grpo-token-for-grad}). 

\small

\begin{align}
\nabla_\theta \mathcal{J}^* = &\; 
\mathbb{E}_{q,\{o_i\}} \Bigg[
\frac{1}{G} \sum_{i=1}^{G^+} \frac{\tilde{A}_i}{|o_i|} \left(
    2 \sum_{t \in c_i} g_{i,t} +
    \sum_{t \in u_i} g_{i,t}
\right) \notag \\
&\qquad + 
\frac{1}{G} \sum_{i=1}^{G^-} \frac{\tilde{A}_i}{|o_i|}
    \sum_{t \in u_i} g_{i,t}
\Bigg]
\end{align}

\normalsize

\noindent where:
\begin{equation}
g_{i,t} = r_{i,t}\,\nabla_\theta \left[\log\big(\pi_\theta(o_{i,t}\mid q, o_{i,<t})\big)\right]
\end{equation}

In addition, following our objective, the adjusted advantage for trajectory~\(i\) becomes:
\begin{equation}
    \tilde{A}_i = A_i - \gamma\,\langle H \rangle_i,
\end{equation}
where \(\langle H \rangle_i\) represents the average policy entropy over all tokens of trajectory \(i\), and \(\gamma\) controls the strength of the entropy correction.

\par \textbf{Reward Settings.}
The reward used in this paper is composed of two components: the first one refers to the format of the model’s answer, and the second one refers to the accuracy of the model’s answer.
The first component assesses formatting: the model receives a score of 10 if it includes all predefined special formatting tokens specified in the prompt (see box \ref{promptbox}). If only a subset of the reasoning or answer tokens are present, the model is awarded 1 point; otherwise, it scores 0. The second component evaluates accuracy: the model obtains a score of 10 if the correct target answer is correctly enclosed within the answer tokens; otherwise, it receives 0. The final reward is the sum of these two components and is provided to guide the learning algorithms.

\begin{tcolorbox}[colback=gray!10!white,colframe=black, title=\ref{promptbox}~\textbf{PROMPT}, label={promptbox}]
You are a helpful assistant for solving math problems. \\
Given a question, first think step by step between 
\texttt{<reasoning>} and \texttt{</reasoning>}. \\
Then, give the final answer between 
\texttt{<answer>} and \texttt{</answer>}.
\end{tcolorbox}




\par \textbf{Training Settings.}
Listing~\ref{lst:training-settings} reports the main training parameters used for GRPO, GTPO and SFT, for both LLaMA and Qwen models. To implement GTPO we used Unsloth~\cite{unsloth}.
\medskip

\begin{lstlisting}[
    style=yaml-box, 
    caption={Training settings for GTPO, GRPO and SFT.}, 
    captionpos=b, 
    label={lst:training-settings}, 
    numbers=none,         % Rimuove i numeri
    xleftmargin=0pt,      % Rimuove lo spostamento a destra (rientro)
    linewidth=\linewidth  % Forza il box ad occupare tutta la larghezza della colonna
]
model_name: "meta-llama/meta-Llama-3.1-8B-Instruct" / "Qwen/Qwen2.5-3B-Instruct"
max_seq_length: 5500
max_prompt_length: 4000
lora_rank: 128 / 64 (Qwen)
load_in_4bit: true
gpu_memory_utilization: 0.4
target_modules:
  - "q_proj"
  - "k_proj"
  - "v_proj"
  - "o_proj"
  - "gate_proj"
  - "up_proj"
  - "down_proj"
random_seed: 3407
warmup_ratio: 0.005
learning_rate: 1e-6
adam_beta1: 0.999999 / 0.9 (GRPO & SFT)
adam_beta2: 0.999999 / 0.95 (GRPO & SFT)
weight_decay: 0.1
beta: 0.0/0.04/10e-6 (GRPO) / "None" (GTPO & SFT)
lr_scheduler_type: "cosine"
optimizer: "paged_adamw_8bit"
logging_steps: 1
per_device_train_batch_size: 1
gradient_accumulation_steps: 1
num_generations: "8/12" / "None" (SFT)
num_train_epochs: 1
num_iterations: 1
save_steps: 500
max_grad_norm: 0.1
report_to: ["wandb"]
\end{lstlisting}

\section{Complementary results}
\label{sec:complementary}
\subsection{Impact of KL $\beta$-coefficient on GRPO.}
\begin{figure*}[t]
    \centering
    \begin{subfigure}{\textwidth}
        \centering
        \includegraphics[width=\textwidth]{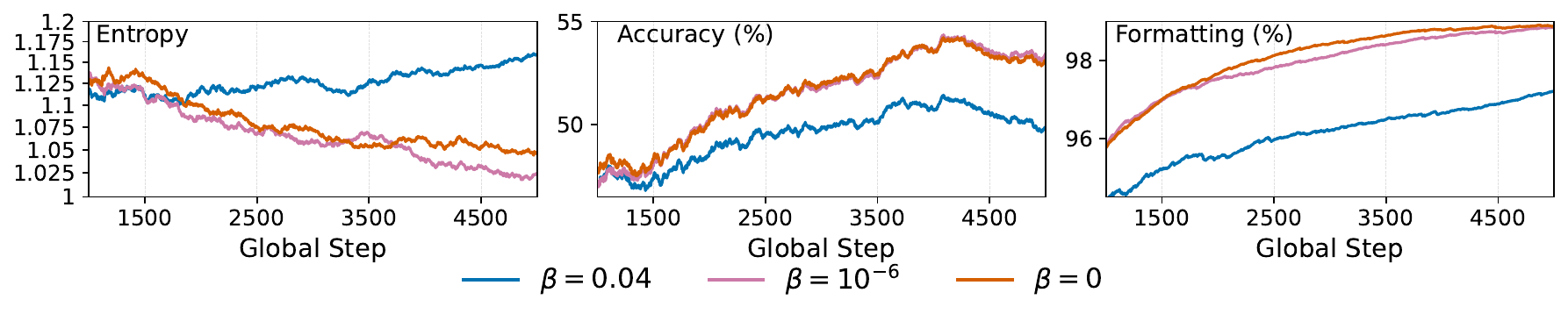}
        \caption{Average entropy (left), accuracy (center), and formatting rate (right) in QWEN trained with GRPO on MATH with different values of \(\beta\) of KL.}
        \label{fig:ablation_math_qwen}
    \end{subfigure}
    
    
    \begin{subfigure}{\textwidth}
        \centering
        \includegraphics[width=\textwidth]{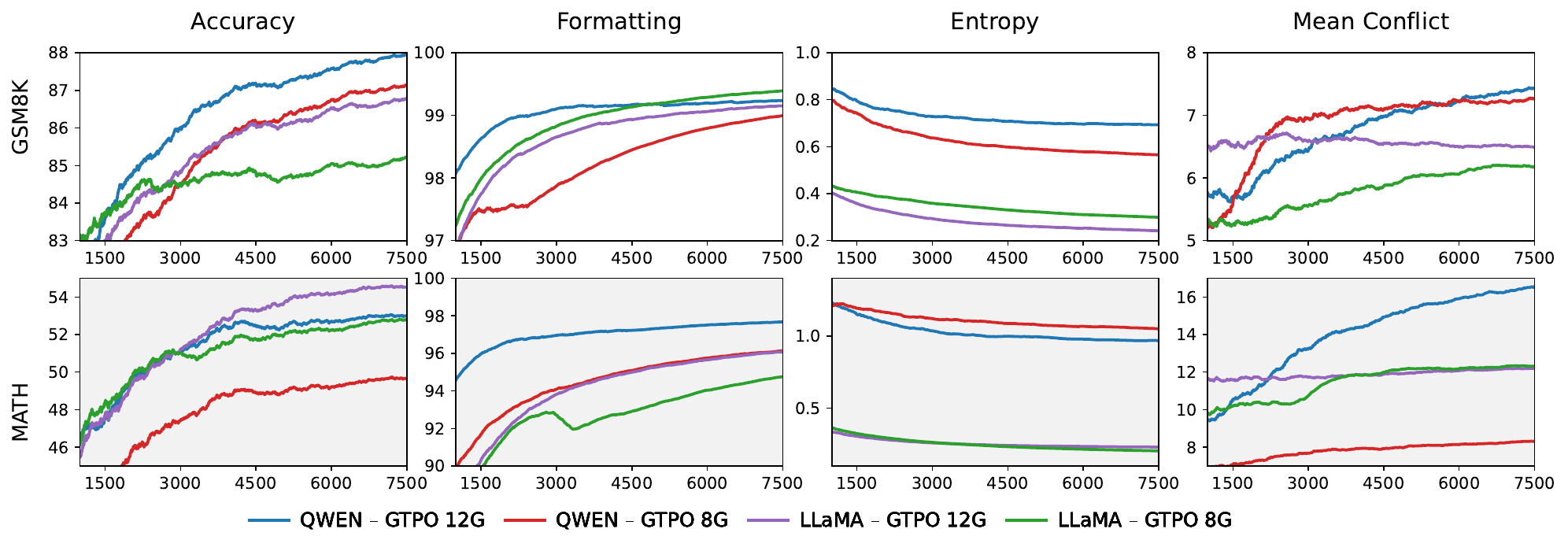}
        \caption{Impact of GTPO group size (\(G=8\) vs. \(G=12\)) on GSM8K (top) and MATH (bottom) for QWEN and LLaMA. Metrics over training steps: accuracy, formatting reward, entropy, and mean conflict (left to right).}
        \label{fig:all_gtpo}
    \end{subfigure}
    
    \caption{Ablation on KL regularization and GTPO group size for QWEN and LLaMA on MATH and GSM8K.}
    \label{fig:combined_math_qwen_gtpo}
\end{figure*}

Figure~\ref{fig:ablation_math_qwen} presents an ablation study on the effect of the KL divergence coefficient \(\beta\) during GRPO training on the MATH dataset using the Qwen model. The three plots respectively show: (left) the average entropy, (center) the accuracy rate, and (right) the formatting rate across training steps.

In the leftmost plot, we observe that with \(\beta = 0.04\) (blue line), the entropy steadily increases, indicating that the model becomes progressively less confident in its predictions. Conversely, smaller or null values of \(\beta\) (orange for \(\beta = 0\) and pink for \(\beta = 10^{-6}\)) result in a more stable or decreasing entropy, reflecting more deterministic behavior during generation.

The central plot highlights how lower values of \(\beta\) lead to better accuracy. In particular, \(\beta = 0\) achieves the highest accuracy throughout training, while \(\beta = 0.04\) results in slower improvement and lower final performance. This suggests that a strong KL regularization term may overly constrain the policy and hinder learning.

Finally, the rightmost plot shows the formatting reward. Both \(\beta = 0\) and \(\beta = 10^{-6}\) achieve high formatting scores (above 98\%) by the end of training, whereas \(\beta = 0.04\) lags behind. This confirms that a weaker KL constraint facilitates the preservation of structural consistency and formatting in model completions. Overall, the figure demonstrates that reducing or removing the KL divergence term in GRPO (\(\beta \rightarrow 0\)) leads to better performance in terms of both accuracy and formatting, while avoiding the entropy inflation observed with larger \(\beta\) values.

\begin{figure}[t]
  \centering
  \includegraphics[width=0.48\textwidth]{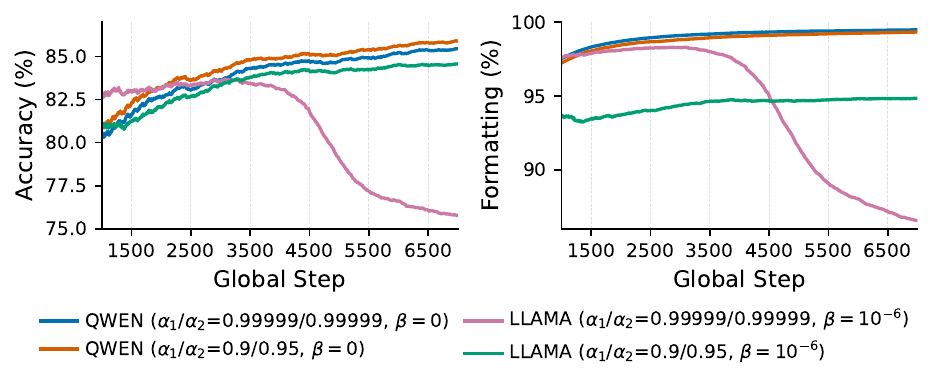}
  \caption{GRPO-$G$=8, comparison between different adam $\alpha_1$ and $\alpha_2$ values.}
  \label{fig:adam_grpo}
\end{figure}

\subsection{Sensitivity of GRPO to Adam Momentum.} Figure~\ref{fig:adam_grpo} compares the impact of different Adam optimizer hyperparameter configurations, specifically the momentum coefficients \(\alpha_1\) and \(\alpha_2\), on the training dynamics of GRPO with generation size \(G = 8\). The two plots report the evolution of accuracy (left) and formatting score (right) over the global training steps. The experiments are conducted using two models, Qwen and LLaMA, each evaluated under two Adam settings:\(\alpha_1 = 0.999999\), \(\alpha_2 = 0.999999\) (used in GTPO~\cite{dohare2023overcoming});\(\alpha_1 = 0.9\), \(\alpha_2 = 0.95\) (the one used by original GRPO~\cite{GRPO}). For Qwen (blue and orange curves), the optimizer settings do not significantly impact training stability: both configurations lead to consistent improvements in accuracy and formatting over time, with the GRPO originals (\(\alpha_1 = 0.9\), \(\alpha_2 = 0.95\)) slightly outperforming in final accuracy. In contrast, for LLaMA (pink and green curves), the choice of \(\alpha_1/\alpha_2\) has a substantial effect. Using the GTPO ones (\(\alpha_1 = 0.99999\), \(\alpha_2 = 0.999999\)) causes a sharp degradation in both accuracy and formatting after approximately 3500 steps, symptomatic of policy collapse. On the other hand, the GRPO default configuration results in more stable formatting behavior and mitigates collapse, although final formatting remains below 95\%.
These results suggest that GRPO training with LLaMA is highly sensitive to Adam hyperparameters, and that careful tuning of \(\alpha_1\) and \(\alpha_2\) is crucial to maintain training stability and prevent collapse, particularly when using smaller values of \(\beta\).

\subsection{Effect of GTPO Group Size on Training Dynamics}

Figure~\ref{fig:all_gtpo} compares GTPO with two group sizes (\(G{=}8\) vs.\ \(G{=}12\)) on
\textbf{GSM8K} and \textbf{MATH} for \textbf{QWEN} and \textbf{LLaMA}, reporting
\emph{accuracy}, \emph{formatting}, \emph{entropy}, and \emph{mean conflict}. On \textbf{GSM8K}, larger groups help both models: \textbf{QWEN 12G} achieves the best final
accuracy, followed by \textbf{LLaMA 12G}, while both 8G settings converge more slowly and to
lower plateaus. All runs exceed \(98\%\) formatting, but larger groups stabilize formatting
earlier, with \textbf{QWEN–12G} reaching the highest levels. LLaMA maintains lower entropy
than QWEN; increasing group size slightly raises entropy for LLaMA but not for QWEN. Mean
token-level conflict increases with group size (12G \(>\) 8G), and QWEN shows higher conflict
than LLaMA, consistent with its higher entropy and trajectory diversity; under GTPO, this
diversity remains beneficial and does not harm formatting. On \textbf{MATH}, the advantage of larger groups persists. \textbf{LLaMA 12G} achieves the
highest accuracy, with \textbf{LLaMA 8G} close behind; QWEN benefits from 12G but remains
below LLaMA. QWEN attains the highest formatting scores, especially with 12G, reaching
near-perfect formatting early in training, while LLaMA also improves with 12G but requires
more steps to reach comparable levels. As on GSM8K, LLaMA stays lower-entropy than QWEN, and
mean conflict grows with group size, most noticeably for \textbf{QWEN 12G}, whereas LLaMA
shows more moderate conflict and weaker sensitivity to group size.
\end{document}